\documentclass{ws-jmrr}
\usepackage{textcomp}
\usepackage{xcolor}
\usepackage[sort,compress,super]{cite}
\usepackage{url}
\usepackage{gensymb}
\usepackage{amsmath,mathtools}
\usepackage{comment}
\usepackage{cite}
\usepackage{amsmath}
\usepackage{amssymb}
\usepackage{xfrac}
\usepackage{multirow}
\usepackage{enumitem}
\usepackage{orcidlink}

\begin{document}

\catchline{0}{0}{2013}{}{}

\markboth{Kheradmand and Yamamoto et al.}{The OncoReach Stylet for Brachytherapy: Design Evaluation and \textcolor{black}{Pilot Study}}

\title{The OncoReach Stylet for Brachytherapy: \\ Design Evaluation and \textcolor{black}{Pilot Study}}

\author{Pejman Kheradmand\orcidlink{0000-0001-8579-1443}\textsuperscript{a*}, Kent K. Yamamoto\orcidlink{0000-0002-9296-8056}\textsuperscript{b,c}\thanks{Equal First Author Contribution}, Emma Webster\orcidlink{0009-0009-8498-2008}$^a$, Keith Sowards\orcidlink{0009-0000-0852-1976}$^d$, \\ Gianna Hatheway\orcidlink{0009-0003-5761-5276}$^e$, Katharine L. Jackson\orcidlink{0000-0001-8574-5869}$^c$, Sabino Zani Jr.\orcidlink{0000-0003-1939-4778}$^{b,c}$, Julie A. Raffi\orcidlink{0009-0006-4164-9094}$^e$, \\ Diandra N. Ayala-Peacock\orcidlink{0000-0002-8032-4499}$^e$, Scott R. Silva\orcidlink{0000-0002-6028-3685}$^d$, Joanna Deaton Bertram\orcidlink{0000-0002-9254-2629}$^b$, and Yash Chitalia\orcidlink{0000-0002-6291-3492}$^a$}

\address{$^a$Healthcare Robotics and Telesurgery (HeaRT) Laboratory, J. B. Speed School of Engineering, University of Louisville, Louisville, Kentucky, USA.\\
E-mail: pejman.kheradmand@louisville.edu}
\address{$^b$Department of Mechanical Engineering and Materials Science, Duke University, Durham, NC, USA.}
\address{$^c$Surgical Education and Activities Lab (SEAL), Duke University Department of Surgery, Durham, NC, USA.}
\address{$^d$ Department of Radiation Oncology, University of Louisville School of Medicine, Louisville, KY, USA.}
\address{$^e$ Department of Radiation Oncology, Duke University School of Medicine, Durham, NC, USA.}

\maketitle

\begin{abstract}
 Cervical cancer accounts for a significant portion of the global cancer burden among women. Interstitial brachytherapy (ISBT) is a standard procedure for treating cervical cancer; it involves placing a \textcolor{black}{radioactive source through a straight hollow needle within or }in close proximity to the tumor and surrounding tissue. However, the use of straight needles limits surgical planning to a linear needle path. We present the OncoReach stylet, a handheld, tendon-driven steerable stylet designed for compatibility with standard ISBT 15- and 13-gauge needles.
 Building upon our prior work, we evaluated design parameters like needle gauge, spherical joint count and spherical joint placement, including an asymmetric disk design to identify a configuration that maximizes bending compliance while retaining axial stiffness. Free space experiments quantified tip deflection across configurations, and a two-tube Cosserat rod model accurately predicted the centerline shape of the needle for most trials. The best performing configuration was integrated into a reusable handheld prototype that enables  manual actuation. A patient-derived, multi-composite phantom model of the uterus and pelvis was developed to conduct a \textcolor{black}{pilot study} of the OncoReach steerable stylet with one expert user. Results showed the ability to steer from less-invasive, medial entry points to reach the lateral-most targets, underscoring the significance of steerable stylets.
\end{abstract}


\begin{multicols}{2}
\section{Introduction}\label{sec:intro}
Cervical cancer incidence is increasing among 30 to 34-year-old women in the United States with an annual percentage change of $2.5\%$ from 2012 to 2019. An estimated 227,062 women were diagnosed with cervix cancer from 2001 to 2019, with approximately 20,697 women between the ages of 30 and 34 years old~[\citen{jama}]. There are expected to be 13,360 additional cases of cervix cancer in the United States in 2025, with an estimated 4,320 deaths~[\citen{Siegel2025}].
Given the rising incidence and mortality rates, there is a growing need for advancements in surgical procedures such as brachytherapy, which is a radiation therapy technique in which a radioactive source is placed within or adjacent to a tumor~[\citen{Brachytherapy_An_overview}].
Accurate needle positioning is essential for achieving highly conformal radiation therapy~[\citen{Imaging_of_implant_needles}]. However, the use of straight needles restricts surgical planning to linear paths, which can be problematic when organs obstruct access to the tumor. These organs, known as organs at risk (OARs), pose significant challenges during needle placement.
In particular, the uterus is anteverted in about $70$–$75\%$  of women, in which the cervix and uterus tilt anteriorly toward the bladder, often complicating the insertion of the tandem (a metal tube inserted through the vaginal opening into the uterine canal) and needles during standard interstitial brachytherapy (ISBT) [\citen{Anteverted_Uterus, Millard2024_qj}]. \textcolor{black}{Typically, these needles are manually inserted through a Syed template, which is a perforated plastic grid fixed to the perineum, to maintain their parallel alignment [\citen{Stock1997}].} These anatomical challenges reduce the accuracy and efficiency of brachytherapy treatment delivery, with the concomitant requirement for more interstitial needles to target the tumor as accurately as possible~[\citen{TAMS2024}].

A clinical case of a markedly anteverted uterus is illustrated in Fig.~\ref{fig:MRI}. This case presents a tumor involving the entire cervix and uterus treated with interstitial brachytherapy needles and a uterine tandem. Fig.~\ref{fig:MRI}(a, b) depicts axial and sagittal views of ISBT using straight needles and straight stylet (a rigid metallic rod to guide needle placement) to treat the tumor. With straight stylet, the brachytherapy plan can safely cover only the medial portion of the tumor (red outline) as shown by the $100\%$ prescription isodose line (yellow outline). The needles cannot advance further to reach the entire tumor due to the risk of puncturing the sigmoid colon. In contrast, Fig.~\ref{fig:MRI}(c, d) showcases the axial and sagittal views, respectively, of a proposed steerable needle plan, highlighting its advantages. The steerable needles can safely advance to the uterine fundus, enabling $100\%$ prescription dose coverage of the anterior, inferior, and lateral portions of the uterus.
\begin{figurehere}
\begin{center}
\centerline{\includegraphics[width=3.3in,keepaspectratio]{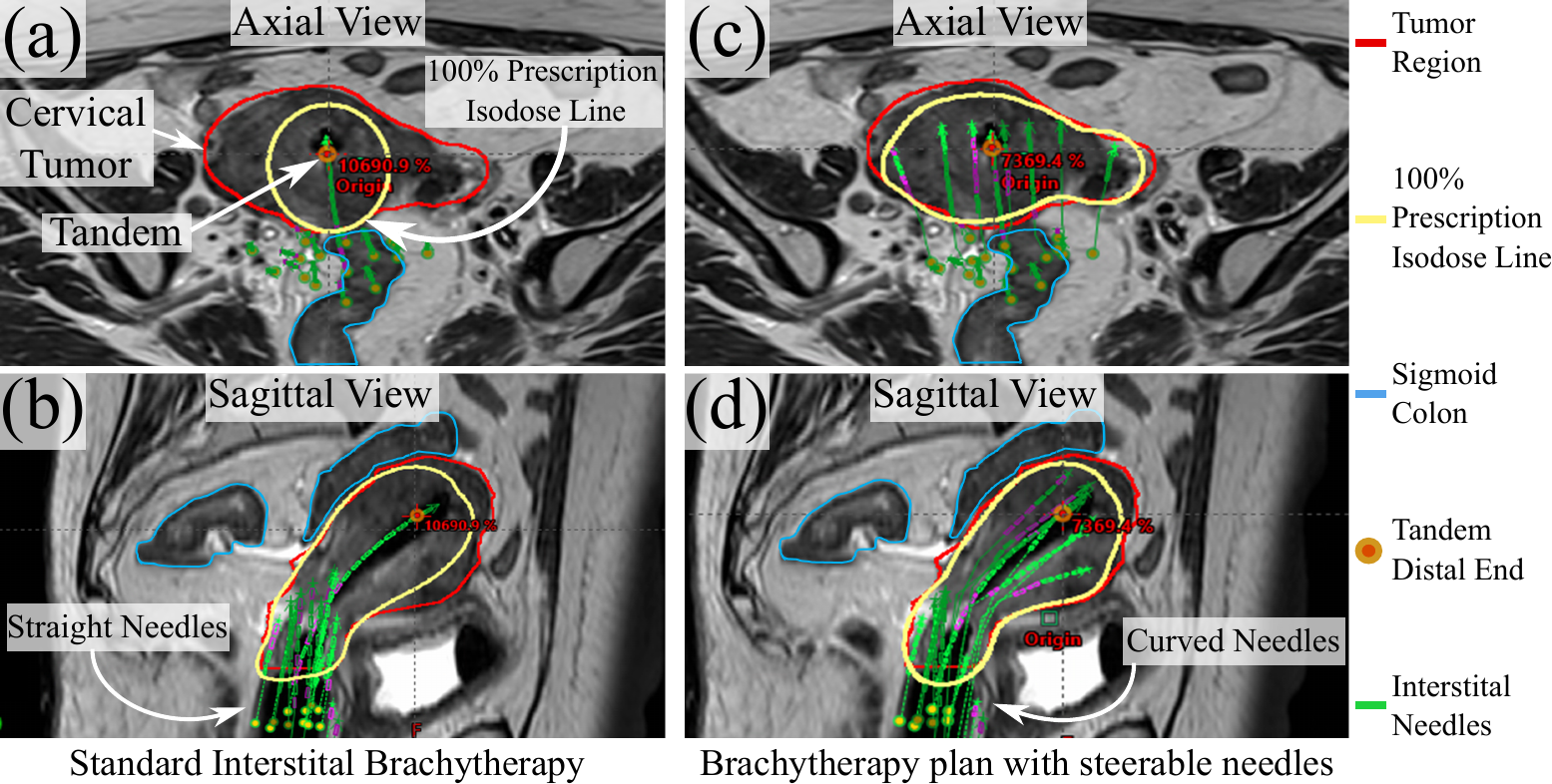}}
\caption{\textcolor{black}{MRI images of a tumor involving the cervix and uterus treated with brachytherapy. (a) Axial view and (b) sagittal view of standard interstitial brachytherapy (ISBT) using straight needles. (c) Axial view and (d) sagittal view of a hypothetical ISBT plan using steerable needles.}}
\label{fig:MRI}
\end{center}
\vspace{-8 mm}
\end{figurehere}

This plan enables greater tumor coverage and, consequently, improved targeting accuracy whilst avoiding puncturing other vital organs in the surrounding area. This approach may also reduce the number of needles required during treatment, potentially decreasing the typical count from 15-20 needles to significantly fewer~[\citen{TAMS2024}].

Previous studies have explored the development of robotic steerable stylets for brachytherapy. Gunderman et al.~[\citen{Gunderman2022}] developed a tendon-actuated deflectable stylet equipped with MR-active tracking. They use agar and porcine tissue samples to test their stylet.
Other designs~[\citen{Deaton2023SteerableStylet, deaton2021robotically, Deaton2023}] have explored tendon-driven continuum robots with a micro-machined stylet. 
However, these designs have limited axial rigidity, which restricts their ability to generate controlled curvature within the tissue.
In our previous work~[\citen{TAMS2024}], we addressed this design limitation by developing a tendon-assisted, magnetically steerable robotic stylet that incorporates magnetic spherical joints and is designed for insertion into 15-gauge ISBT plastic needles.
We achieved significant tip deflection both in free space and within gelatin-based phantom tissue.
Although our previous work demonstrated that dual actuation (magnetic plus tendon) produced greater tip deflection in free space compared to single-mode actuation, the in-vitro experiments using phantom tissue with stiffness comparable to human tissue showed that the curvature achieved with dual actuation was comparable to that obtained with tendon-only actuation (see Fig.~\ref{fig:Vitro_Bertram}). 

\begin{figurehere}
\begin{center}
\centerline{\includegraphics[width=3.1in,keepaspectratio]{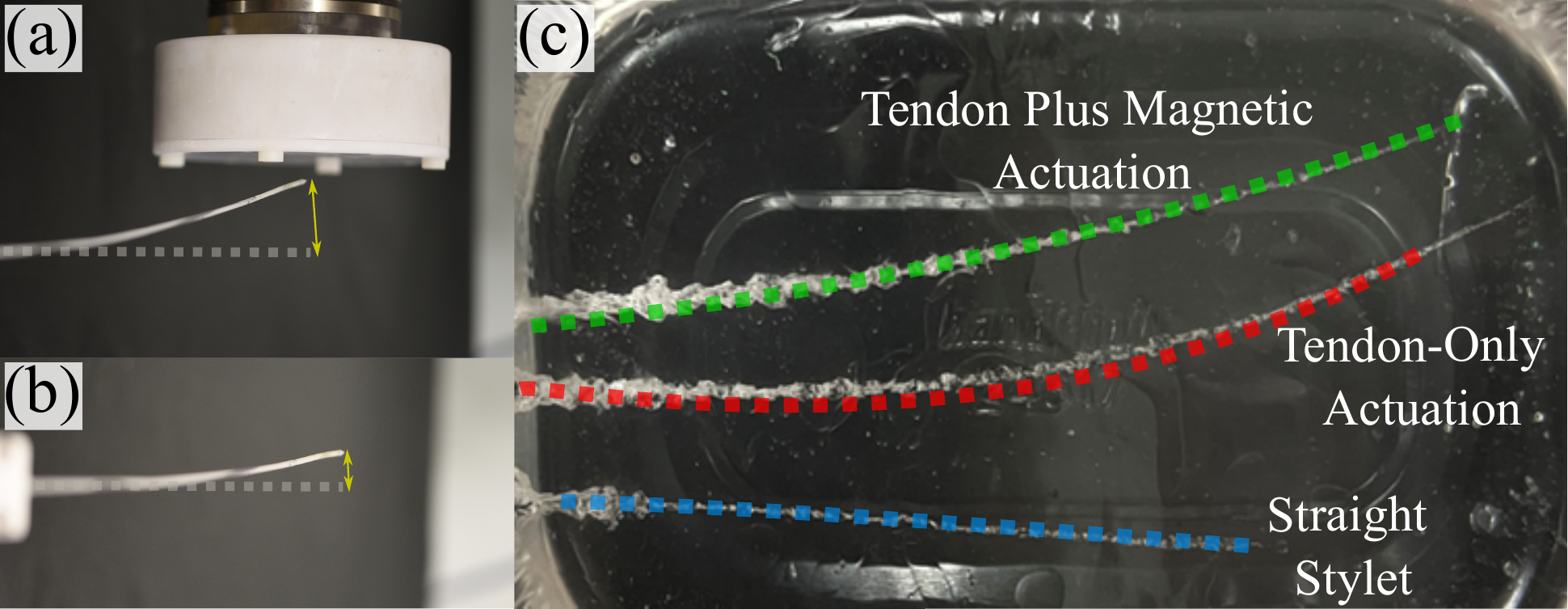}}
\caption{Comparison of a steerable needle configurations under different actuation modes~[\citen{TAMS2024}]. (a) Dual actuation using tendon and magnets and (b) tendon-only actuation. (c) Comparison of both methods to straight stylets.}
\label{fig:Vitro_Bertram}
\end{center}
\end{figurehere}
\vspace{-8 mm}

In this work, we developed a non-magnetic, tendon-driven steerable stylet with a handheld controller \textcolor{black}{(Section~\ref{sec:Design})}. Multiple design parameters are systematically evaluated and validated with a proposed mechanical model \textcolor{black}{(Section~\ref{sec:modeling})} to identify the most effective configuration. \textcolor{black}{In parallel, a high-fidelity uterus-and-pelvis phantom model was developed to emulate the mechanical stiffness comparable to that of human tissue (Section~\ref{sec:phantomdevelopment}). Free-space testing of the stylet designs are conducted, and the best-performing design is integrated into a novel handheld prototype (Section~\ref{sec:Experiments}). The handheld design is then validated with an expert user who inserted the OncoReach stylet into the uterus-and-pelvis phantom. Finally, Section~\ref{sec:conclusion} summarizes findings and discusses future work.} 

\section{Design}\label{sec:Design}
\vspace{-3 mm}
\subsection{Front-end and Back-end Design}\label{subsec:FrontEnd_design}
Fig.~\ref{fig:JointsDesign}(a) illustrates the overall setup of the system.  The front-end consists of the needle-stylet assembly, in which the stylet is inserted coaxially into a commercially available brachytherapy needle (Best Medical International, Inc., Springfield, VA, USA).
Fig.~\ref{fig:JointsDesign}(b) illustrates the OncoReach stylet, which consists of a nickel-titanium (nitinol) tube integrated with connected spherical joints separated by 3D-printed routing disks. Each disk incorporates four channels to guide the tendons. By selectively actuating individual tendons, the stylet bends toward the corresponding direction of the applied tendon force. Multiple variations of the routing disks are designed for this study.
Fig.~\ref{fig:JointsDesign}(c) presents a $1.4$~mm diameter disk compatible with 15-gauge needles, while Fig.~\ref{fig:JointsDesign}(d) shows the cross-sectional view of a same-sized disk used at the nitinol tube interface. Since the tube’s inner diameter is smaller than the routing path inside the 
\begin{figurehere}
\begin{center}
\centerline{\includegraphics[width=2.7 in,keepaspectratio]{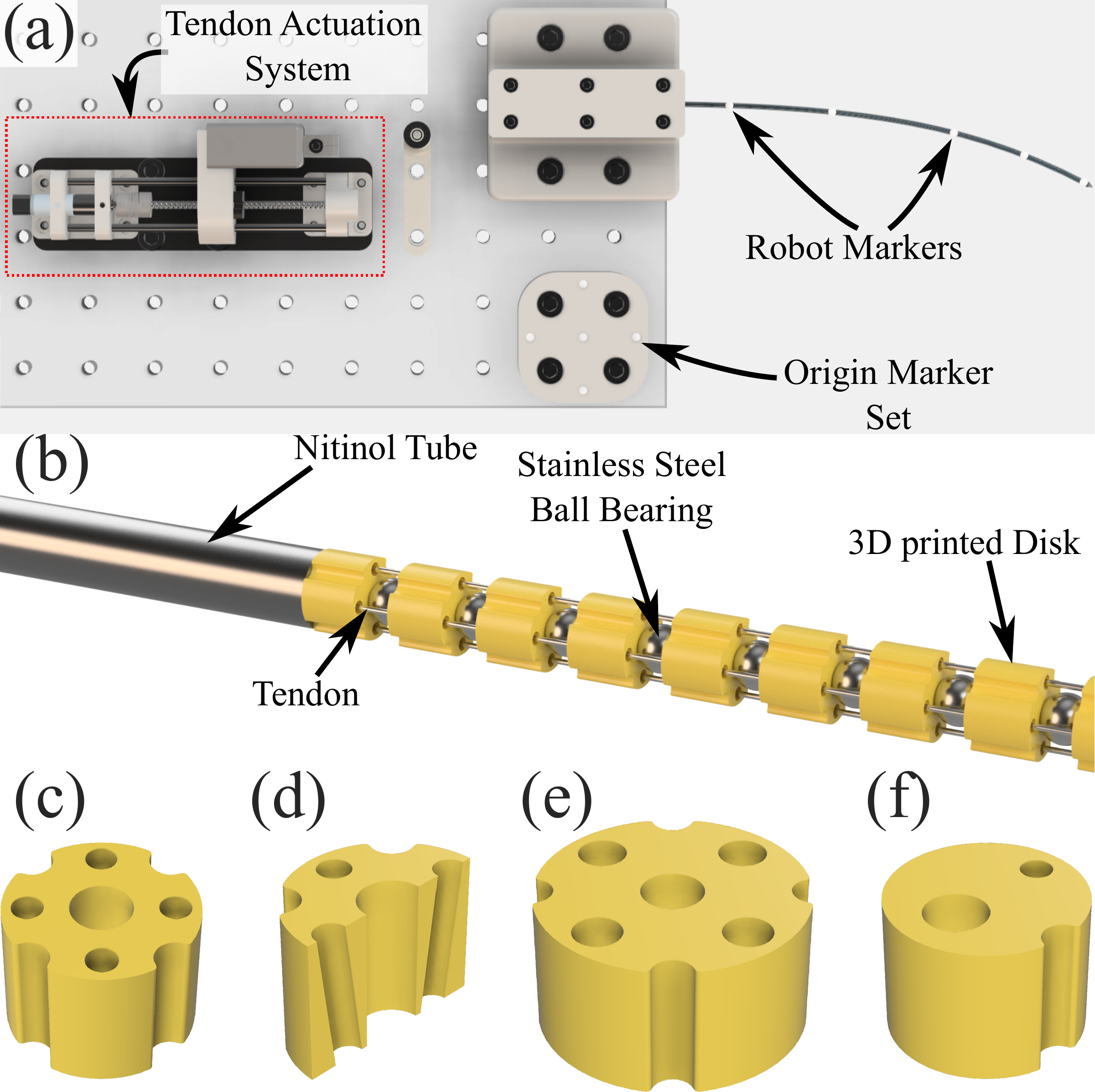}}
\caption{OncoReach stylet design. \textcolor{black}{(a) System overview showing the tendon-actuation system, needle-stylet, and tracking markers for shape measurement.} (b) Stylet design. (c) Routing disks for 15 gauge needle. (d) Cross-section of tilted tendon channels for smooth routing. (e) Routing disks for 13 gauge needles. (f) Asymmetric single-tendon disk with offset ball bearing for enhanced bending.}
\label{fig:JointsDesign}
\end{center}
\vspace{-7 mm}
\end{figurehere}
disk, the tendon channels are inclined to ensure smooth tendon passage between the tube and disk assembly. Fig.~\ref{fig:JointsDesign}(e) shows a $1.95$~mm version designed for 13-gauge needles. 
The corresponding nitinol tubes have inner and outer radii of $R_\text{in}=0.533$~mm and $R_\text{out}=0.673$~mm for the 15-gauge tube, and $R_\text{in}=0.747$~mm and $R_\text{out}=0.965$~mm for the 13-gauge tube, as specified by the manufacturer.
Both configurations in Fig.~\ref{fig:JointsDesign}(c) and ~\ref{fig:JointsDesign}(e) employ a $0.8$~mm stainless-steel ball bearing at the center, functioning as a spherical joint. In contrast, Fig.~\ref{fig:JointsDesign}(f) depicts an asymmetric, single-tendon-routed disk in which the ball bearing is offset to one side and the tendon channel is positioned on the opposite side. This arrangement increases the moment arm generated by the tendon force, thereby enhancing bending compliance. 
Additionally, we investigate the effects of joint placement along the stylet body. Three configurations are tested, with the spherical joints positioned near the base, midsection, and distal end of the stylet. Finally, to study the influence of joint count, we test stylet prototypes with 15, 20, and 30 spherical joints.

At the back-end, an $\phi8$~mm DC motor (Maxon Metal Brushes, 0.5~W, Maxon international ltd., Switzerland) coupled to a lead screw converts rotary motion into linear displacement. The motion drives a $5$~lb load cell (MDB-5, Transducer Techniques) for real-time tendon tension measurement. A nitinol tendon is fixed to the load cell, allowing precise tension control via a closed-loop PID controller implemented in MATLAB/Simulink using a Texas Instruments LAUNCHXL-F28379D LaunchPad.
\vspace{-4 mm}
\subsection{Handheld Prototype Design}\label{subsec:Handheld_design}
\vspace{-2 mm}

\textcolor{black}{To enhance clinical usability, a handheld, single-direction prototype was then developed (see Fig.~\ref{fig:HandHeldDesign}). This design allows for manual insertion and steering by using a finger grip to actuate the tendon directly. The grip slides on a $3$~mm guide shaft supported by linear ball bearings to ensure smooth motion. Pulling the grip controls stylet bending via a connector at the shaft's distal end, providing the operator with intuitive manual control during tissue insertion. Because the bending direction is constant, the user must rotate his or her wrist to adjust the steering orientation.}

\begin{figurehere}
\begin{center}
\centerline{\includegraphics[width=2.9in,keepaspectratio]{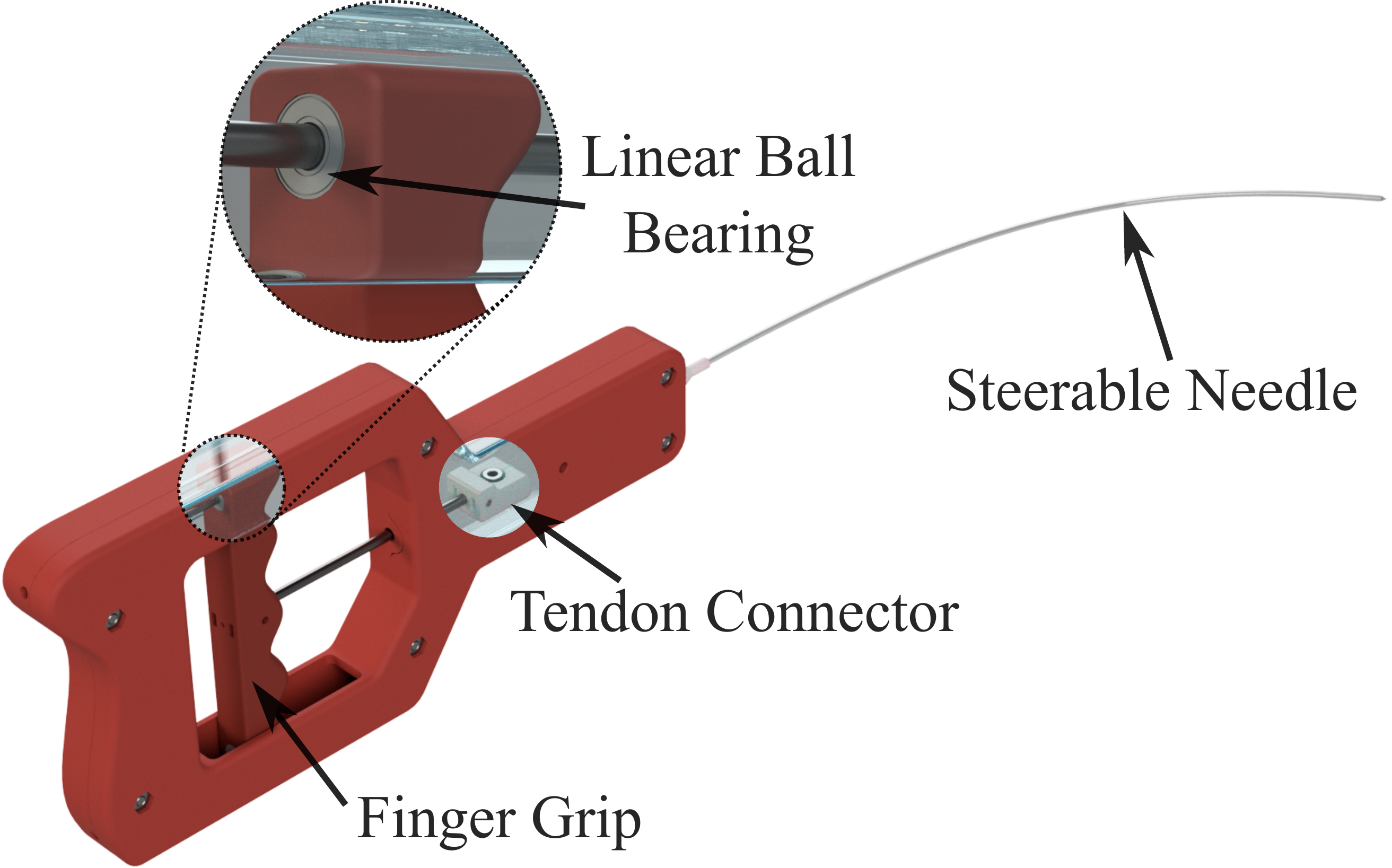}}
\caption{Handheld prototype design of the OncoReach stylet and needle.}
\label{fig:HandHeldDesign}
\end{center}
\vspace{-8 mm}
\end{figurehere}

\vspace{-5 mm}
\section{Mechanical Model}\label{sec:modeling}
\vspace{-1 mm}
In this section, we present a two-tube Cosserat rod model to predict the shape of the system in free space under applied tendon tension. 
In contrast to~[\citen{TAMS2024}], the model proposed here provides a more accurate representation of the relative shear and axial dilation along the shared centerline.
\vspace{-2 mm}
\subsection{Static Model}\label{subsec:Static_Model}
\vspace{-3 mm}
The proposed model incorporates both the stylet and the needle: the stylet, referred to as tube~$1$ (inner tube), and the needle, referred to as tube~$2$ (outer tube). Both tubes share a common centerline but can twist and elongate/contract independently along their lengths. Figure~\ref{fig:Model} depicts the system design parameters and the local material frames associated with each tube. Following the convention of~[\citen{CalebStatic2011}], the differential kinematics of the shared centerline $\bm{p}_c(s) \in \mathbb{R}^3$, parameterized by the arc length $s$, are expressed as:
\begin{align}\label{eq:shape_evol}
    \dot{\bm{p}}_c(s) = R_i(s)\bm{v}_i(s), \qquad 
    \dot{R}_i(s) = R_i(s)[\bm{u}_i(s)]
\end{align}
The vectors $\bm{v}_i(s)$ and $\bm{u}_i(s)$ represent the linear strain rate and angular strain rate, respectively.
The material frame $R_i(s) \in \mathrm{SO}(3)$ of tube $i\in [1,2]$ is defined along the arc length $s$, where $s \in [0, l]$ and $l$ denotes the total length of the tube. The operator $[\,\cdot\,]$ converts a vector in $\mathbb{R}^3$ to its skew-symmetric form in $\mathfrak{so}(3)$ (the Lie algebra of $\mathrm{SO}(3)$). Derivatives with respect to arc length are denoted by $\dot{q} = \sfrac{dq}{ds}$. Dependence on $s$ is omitted for brevity in subsequent derivations. Each tube’s material frame is defined by its orthonormal directors $[\bm{d}_{1,i}, \bm{d}_{2,i}, \bm{d}_{3,i}]$, which form the 
\begin{figurehere}
\begin{center}
\centerline{\includegraphics[width=2.5in,keepaspectratio]{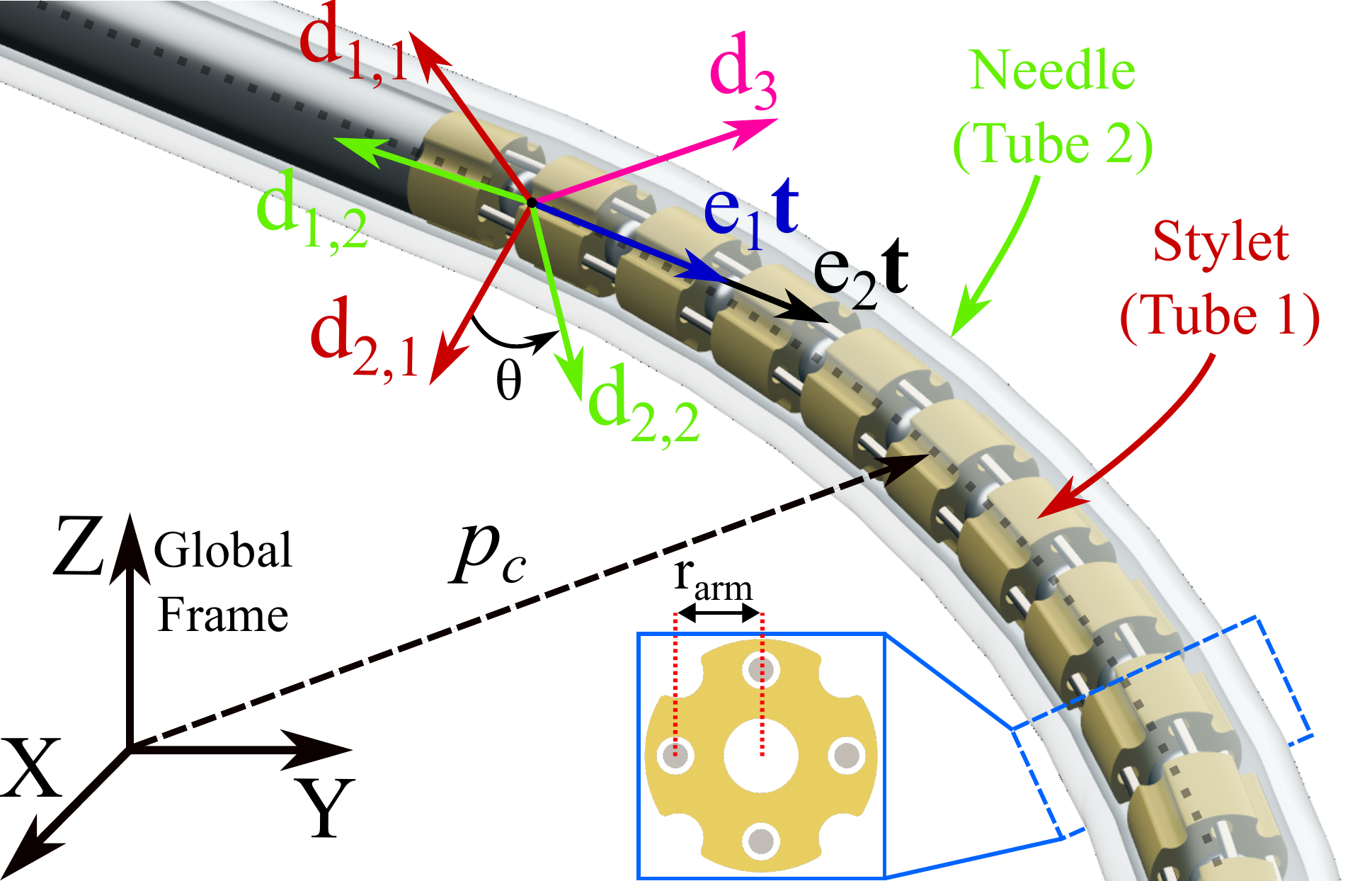}}
\caption{Schematic of the OncoReach stylet mechanical model, highlighting the relative twist ($\theta$) and axial dilation ($e_2$) of the outer tube with respect to the inner tube.}
\label{fig:Model}
\end{center}
\vspace{-5 mm}
\end{figurehere}
columns of $R_i$. The third director $\bm{d}_3$, normal to the cross-section, is common to both tubes, i.e., $\bm{d}_3 = \bm{d}_{3,i}$ for all $i\in [1,2]$. The relative rotation between tube~2 and the reference tube~1 is represented by $R_2 = R_1R_{d_3}(\theta)$, where $R_{d_3}(\theta)$ denotes a rotation about $\bm{d}_3$ by angle $\theta$, defining the twist between tubes. In addition to relative twisting, the tubes experience axial dilation during actuation. The dilation factor of tube~2, denoted as $e_2$, is defined relative to tube~1 as $\beta = e_2/e_1$.

For a Cosserat rod subjected to distributed force $\bm{f}_i \in \mathbb{R}^3$ and moment $\bm{\tau}_i \in \mathbb{R}^3$ per unit length, the static equilibrium equations are given by~[\citen{dupont2009design}]:
\begin{align}\label{eq:equilibrium_eq_single_tube}
\begin{bmatrix}
    \dot{\bm{n}}_i \\[3pt]
    \dot{\bm{m}}_i
\end{bmatrix}
+
\begin{bmatrix}
    [\bm{u}_i] & \mathbf{0} \\[3pt]
    [\bm{v}_i] & [\bm{u}_i]
\end{bmatrix}
\begin{bmatrix}
    \bm{n}_i \\[3pt]
    \bm{m}_i
\end{bmatrix}
+
\begin{bmatrix}
    \bm{f}_i \\[3pt]
    \bm{\tau}_i
\end{bmatrix}
= \mathbf{0}
\end{align}
Assuming linear elasticity, the internal force and moment are related to the strain variables through the constitutive model:
\begin{align}\label{eq:constitutive_short}
\begin{bmatrix}\bm{n}_i\\ \bm{m}_i\end{bmatrix}=
\begin{bmatrix}K_{se,i} & \mathbf{0}\\ \mathbf{0} & K_{bt,i}\end{bmatrix}
\begin{bmatrix}\bm{v}_i-\bm{v}^*_i\\ \bm{u}_i-\bm{u}^*_i\end{bmatrix}
\end{align}
where $K_{se,i}=\mathrm{diag}(GA_i,GA_i,EA_i)$ and $K_{bt,i}=\mathrm{diag}(EI_{d_{1},i},EI_{d_{2},i},GJ_{d_{3},i})$ are stiffness matrix for shear and axial elongation, and stiffness matrix for bending and torsion, respectively. $A_i$ is the cross-sectional area of the tube, $E$ is Young's modulus, $G$ is the shear modulus. $I_{d_{1}}$ and $I_{d_{2}}$ represent the second moments of area of the tube's cross section about the principal axes, while $J_{d_{3},i}$ denotes the polar moment of inertia of the cross section about the $d_3$ axis. Also, variables marked with a superscript $^*$ denote their values in the undeformed configuration.
In this study, all distributed forces and moments originate from tendon actuation. The position of the $j$th tendon in the local body frame is denoted by $\bm{r}_{arm,j} = [x_j(s), y_j(s), 0]^T$, where $j$ corresponds to the tendon index (e.g., $j \in [1,4]$ in Fig.~\ref{fig:Model}). Assuming these local coordinates remain constant along the arc length $s$, the distributed tendon-induced force and moment are expressed as:
\begin{align}\label{eq:tendon_forces}
        f_{j} = - \lambda_{j}\frac{[\dot{p}_{j}]^2}{||\dot{p}_{j}||^3}\ddot{p}_{j},
        \qquad
    \bm{\tau}_j = -[\bm{r}_{arm,j}]\,\bm{f}_j
\end{align}
where $\lambda_j$ is the tension applied to the $j$th tendon, and $\bm{p}_j$ denotes its path in the global frame. The first and second derivatives of $\bm{p}_j$ in the body frame are:
\small
\begin{align}\label{eq:tendon_position_derivatives}
    \dot{\bm{p}}_j^b = [\bm{u}_1]\bm{r}_{arm,j} + \bm{v}_1, \qquad
    \ddot{\bm{p}}_j^b = [\bm{u}_1]\dot{\bm{p}}_j^b - [\bm{r}_{arm,j}]\dot{\bm{u}}_1 + \dot{\bm{v}}_1
\end{align}
\normalsize

Because the tubes share a common backbone, their curvatures along $\bm{d}_1$ and $\bm{d}_2$ are equal, while independent twisting is permitted along $\bm{d}_3$. The relative twist rate between the tubes is:
\begin{align}\label{eq:twist_relation}
    \dot{\theta} = u_{2,d_3} - u_{1,d_3}, \qquad 
    \ddot{\theta} = \dot{u}_{2,d_3} - \dot{u}_{1,d_3}
\end{align}

Accordingly, the angular strain of tube~2 can be expressed as:
\begin{align}\label{eq:u2_u1_relationship}
    \bm{u}_2 = R_{d_3}^T(\theta)\bm{u}_1 + \dot{\theta}\,\mathbb{I}_3
\end{align}
where $\mathbb{I}_3 = [0, 0, 1]^T$ represents the unit vector along $\bm{d}_3$. Differentiating Eq.~\eqref{eq:u2_u1_relationship} yields:
\begin{align}\label{eq:u2_dot}
    \dot{\bm{u}}_2 = (R_{d_3}^T(\theta) - \mathbb{I}_{(3,3)})\dot{\bm{u}}_1 + \dot{\theta}\,[\mathbb{I}_3]^T R_{d_3}^T(\theta)\bm{u}_1 + \mathbb{I}_3\,\dot{u}_{2,d_3}
\end{align}
where $\mathbb{I}_{(3,3)} = \text{diag}(\mathbb{I}_3)$. Following the convention in~[\citen{gazzola2018forward}], the dilation factor $e_i = ds_i/ds_i^*$, defined as the ratio between deformed and undeformed arc elements, relates the strain rates of the two tubes as:
\begin{align}\label{eq:shear_relation}
    \bm{v}_2 = \beta R_{d_3}^T(\theta)\bm{v}_1
\end{align}
where $\beta = e_2/e_1$. Differentiating Eq.~\eqref{eq:shear_relation} gives:
\begin{align}\label{eq:shear_dot_relation}
    \dot{\bm{v}}_2 = \dot{\beta} R_{d_3}^T(\theta)\bm{v}_1 + \beta [\mathbb{I}_3]^T R_{d_3}^T(\theta)\dot{\theta}\bm{v}_1 + \beta R_{d_3}^T(\theta)\dot{\bm{v}}_1
\end{align}

This formulation differs from the derivation in~[\citen{TAMS2024}], where it was assumed that a shared centerline implied that $\bm{v}_2$ and $\bm{v}_1$ were directly related by a rotation about $\bm{d}_3$. In contrast, the current formulation correctly distinguishes between linear and bending strain coupling across tubes. Finally, both $\bm{u}_2$ and $\bm{v}_2$, along with their derivatives, are represented in terms of the strain variables of the reference tube ($\bm{u}_1$, $\bm{v}_1$) and their respective derivatives. For concentric tubes, static equilibrium of internal moments and forces in the $\bm{d}_1$–$\bm{d}_2$ plane is expressed as:
\begin{align}\label{eq:moment_sum_local1}
    \sum_{i=1}^2 R_{d_3}(\theta)\big(\dot{\bm{m}}_i + [\bm{u}_i]\bm{m}_i + [\bm{v}_i]\bm{n}_i + \bm{\tau}_i\big) = \bm{0}\big|_{\bm{d}_1,\bm{d}_2} \\[3pt]
\label{eq:force_sum_local1}
    \sum_{i=1}^2 R_{d_3}(\theta)\big(\dot{\bm{n}}_i + [\bm{u}_i]\bm{n}_i + \bm{f}_i\big) = \bm{0}\big|_{\bm{d}_1,\bm{d}_2}
\end{align}
Moreover, for each tube, Eq.~(\ref{eq:equilibrium_eq_single_tube}) is applied along the $\bm{d}_3$ direction. Following the same reasoning as in~[\citen{chitalia2023model}], the equilibrium configuration of the system is determined by combining the in-plane ($\bm{d}_1$–$\bm{d}_2$) force and moment balances with Eq.~(\ref{eq:equilibrium_eq_single_tube}) applied along the $\bm{d}_3$ direction for each tube, together with the tendon-induced loads defined in~(\ref{eq:tendon_forces})–(\ref{eq:tendon_position_derivatives}). The resulting set of coupled differential equations can be expressed compactly as:
\begin{align}\label{eq:Ax_eq_b_short}
A(\mathbf{x})\,\dot{\mathbf{x}} = \bm{b}(\mathbf{x})
\end{align}
the state vector $\mathbf{x} = \{\bm{u}_1, \bm{v}_1, {u}_{2,d_3}, \beta\}$ represents the deformation variables of a system composed of two tubes. Tube $1$ can include up to four tendons, while tube $2$ includes none, as the needle does not incorporate any tendons. 

\vspace{-3 mm}
\section{Uterus-Pelvis Phantom Model}\label{sec:phantomdevelopment}
\vspace{-3 mm}

\subsection{Uterus Phantom Model}
\vspace{-2 mm}
\begin{figurehere}
\begin{center}
\centerline{\includegraphics[width=3.0in,keepaspectratio]{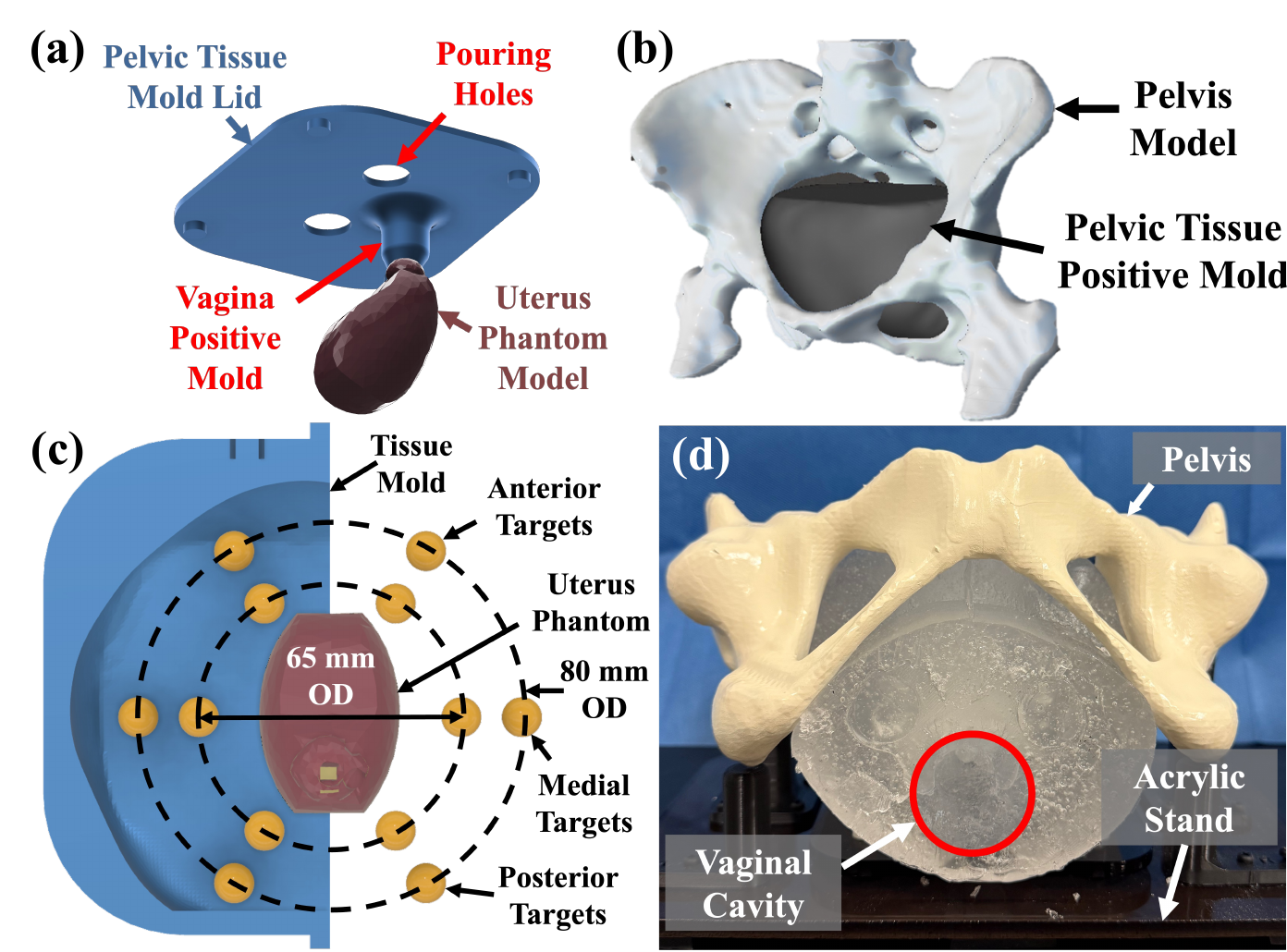}}
\caption{(a) Phantom uterus in CAD with vagina positive mold. (b) 3D model of CT-segmented pelvis with positive mold for pelvic tissue. (c) Top view of pelvic tissue mold with target positions. (d) Assembled phantom model (transverse view).}
\label{fig:phantom}
\end{center}
\vspace{-5 mm}
\end{figurehere}

The uterus phantom model (Fig. \ref{fig:phantom}(a)) uses the methodology from \textcolor{black}{[\citen{julie,model2}]}, in which negative molds of the uterus (average dimensions derived from a 50-patient data study [\citen{model1}]) are PLA 3D printed (Ultimaker S3, Utrecht, Netherlands) for pouring of the hollow uterine shell.
\vspace{-5 mm}
\subsection{Pelvis Phantom Model}
\vspace{-2 mm}
The pelvis model, shown in Fig. \ref{fig:phantom}(b), is derived from a de-identified patient CT scan in which medical imaging software (3DSlicer [\citen{3dslicer}]) is used to segment the pelvis using a Hounsfield Unit (HU) thresholding technique, followed by manual segmentation. The 3D volume is then exported as a stereolithogrpahy (STL) file and imported into 3D mesh modeling software (Meshmixer), in which any residual noise is removed. The pelvis is further modified in Meshmixer to contain only the pubic symphysis, ischiopubic ramus, and illium. The updated STL file is then uploaded to CAD software to create holes for standoffs in the phantom model. The finalized pelvis model is then resin 3D-printed (Formlabs 3BL, Somerville, MA, USA) with clear resin and spray-painted white (Rust-Oleum, Vemon Hills, IL, USA).
\vspace{-3 mm}
\subsection{Pelvic Tissue Phantom Model}
\vspace{-2 mm}
The pelvic tissue comprises of tissue simulation gel (SimuGel \textcolor{black}{\#}3, Humimic, Greenville, SC, USA) with a Young's modulus of 0.19 MPa, density of \textcolor{black}{856.8} kg/m$^3$, and speed of sound of \textcolor{black}{1459} m/s, which \textcolor{black}{have been obtained from the manufacturer [\citen{humimic}]}. The geometry of the tissue is arbitrarily determined by importing a sphere mesh appended inside the pelvic cavity of the modified pelvis model, as depicted in Fig. \ref{fig:phantom}(b). The sphere mesh is then shaped until a tissue volume with no intersections into the pelvis is formed. The tissue model is then imported into CAD software to create a mold. Insertion holes for 10 mm outer diameter (OD) target molds are designed into the interior shell of the mold, positioned as shown in Fig. \ref{fig:phantom}(c). 15 mm tall targets are placed 60$\degree$ apart along a circular path with outer diameter (OD) of 65 mm. Similarly, 5 mm tall targets are placed along an 80 mm OD circle. In addition, an extrusion of 35 mm on the inferior side of the mold is made to represent the tissue surrounding the vaginal canal. The top lid of the mold has a built-in positive mold to create a negative space for the vaginal canal with dimensions designed to accommodate the insertion of the vaginal obturator, as well as an extrusion to hold the uterus model in free space while the Humimic gel cures (Fig. \ref{fig:phantom}(a)). The mold is then split into two parts and printed (Bambu X1C, Shenzhen, China) with screw holes for locking and securing.
\vspace{-5 mm}
\subsection{Model Formation and Assembly}
\vspace{-2 mm}
The uterus model is formed by first casting silicone (Smooth-on Mold Star Silicone 20T, Easton, PA, USA) inside the mold. Once an exterior shell of the silicone uterus is cured, the uterus is cut to remove the internal portion of the PLA mold, and Humimic gel is melted into a liquid form at approximately 200$\degree$C and poured into the hollow uterine shell. Once cast, the phantom is set to rest until cured. Silicone is then cast over the cured Humimic gel to seal off the casting entry point of the uterus.

Once the uterus model is complete, the curved extrusion on the top lid of the pelvic tissue mold is inserted into the Humimic region of the uterus. The target molds are placed in their respective slots, and mold release (Smooth-On Universal Mold Release, Macungie, PA, USA) is sprayed on the interior of the pelvic tissue mold. The pelvic tissue mold is assembled, and Humimic gel solution is then poured into the mold and set to rest for three hours. The pelvic tissue mold is then disassembled, and the Humimic gel with embedded uterus is removed from the mold. The resulting phantom model has a vaginal canal present in the Humimic tissue which connects to the internal silicone uterus (Fig. \ref{fig:phantom}(d)). Stock acrylic board (McMaster-Carr, Elmhurst, IL, USA) is then laser-cut to serve as the base, shown in Fig. \ref{fig:phantom}(d). 3D-printed stands for the pelvic tissue model and ischium are secured onto the acrylic board. The pelvic tissue is placed on the stand, then the pelvis is placed above to secure the pelvic tissue.

\vspace{-4 mm}
\section{Experiments and Results}\label{sec:Experiments}
\vspace{-2 mm}
A series of experiments are conducted to identify the best-performing OncoReach stylet configuration that achieves high bending compliance while maintaining sufficient axial stiffness. Trials are performed by varying the needle and stylet gauges, the number of spherical joints (15, 20, and 30), and the joint placement along the stylet (base, midsection, and tip) \textcolor{black}{in free-space}. An additional asymmetric single-tendon-routed design is also evaluated. The mechanical model described in Section~\ref{sec:modeling} is validated experimentally, and a handheld trial is subsequently performed. Finally, OncoReach stylet is deployed by an expert user in the phantom model for a \textcolor{black}{pilot study} of the tool.
\vspace{-3 mm}
\begin{tablehere}
\tbl{Maximum tip deflection across OncoReach stylet configurations."ga" indicates needle gauge, "Joints" the number of spherical joints, and parentheses denote joint placement.
\label{tab:TipDeflection}}
{\begin{tabular}{@{}lcccr@{}}
\toprule
\multirow{2}{*}{\textbf{Configuration}} & \textbf{Tendon}       & \textbf{Tip}\\
  & \textbf{Tension (N)}  & \textbf{Deflection (mm)}\\ 
\colrule
15~ga.-20~Joints (Tip)   & 9.9  & 9.87\\
13~ga.-20~Joints (Tip)   & 14.0 & 6.49\\
15~ga.-15~Joints (Tip)   & 10.0 & 8.98\\
15~ga.-30~Joints (Tip)   & 9.9  & 11.05\\
15~ga.-20~Joints (Base)  & 9.9  & 29.40\\
15~ga.-20~Joints (Mid)   & 10.0 & 20.00\\
15~ga.-20~Joints (Asym.) & 9.9  & 11.34\\
\botrule
\end{tabular}}
\vspace{-4 mm}
\end{tablehere}

\textcolor{black}{While base and midsection joint placements produce greater displacement, they are incompatible with the Syed template (see Section \ref{sec:intro}). Additionally, the asymmetric configuration was structurally unstable and required excessive reassembly. Therefore, the 15-gauge configuration with 30 joints at the tip was selected for the pilot study.}

\vspace{-4 mm}
\subsection{Free-Space Experimental Setup}\label{subsec:FreeSpaceSetup}
\vspace{-2 mm}
The experimental setup, shown in Fig.~\ref{fig:JointsDesign}(a), is used for all free-space trials. Five reflective markers are placed along each needle and tracked using four Vero~v2.2 motion-capture cameras (Vicon Motion Systems Ltd., United Kingdom). An additional set of five markers at the robot base define a global reference frame for the system. Table~\ref{tab:TipDeflection} summarizes the maximum tip deflection measured for each configuration.
As shown in Table~\ref{tab:TipDeflection}, the 15~gauge configuration exhibits greater bending compliance than the 13~gauge design, even though the latter is actuated with a higher tendon tension ($14$~N compared to $9.9$~N). Although the needle cross-sectional areas are nearly identical, the $13$~gauge stylet has a cross-sectional area $2.21$~times larger than that of the 15~gauge stylet. Consequently, the second moment of area ($I$) increases by approximately $4.47$~times, resulting in greater axial stiffness and reduced bending compliance across the system. The configuration with $30$ spherical joints demonstrates greater bending compliance than those with $15$ or $20$ joints. This behavior arises because the articulated spherical joint section of the stylet has lower bending stiffness than the nitinol stem. The experimentally measured bending stiffness matrix for the spherical joint section is $K_{bt,\text{joints}} = \mathrm{diag}([0.0008,~0.0008,~0.0047])$, \textcolor{black}{which remained constant across all trials. Conversely,} the bending stiffness matrix for the nitinol tube section, calculated from the tube geometry and material properties provided by the manufacturer, is $K_{bt,\text{tube}} = \mathrm{diag}([0.0062,~0.0062,~0.0047])$. Therefore, increasing the length of spherical joints improves bending compliance but can result in buckling during insertion through the Syed template, a planar template with evenly-spaced  needle holes. Placing the low-stiffness joint section near the base produced the highest tip deflection, followed by the midsection, while the tip configuration showed the least. The asymmetric design, with a larger $r_{arm}$, generated greater tip deflection than the symmetric configuration.

\vspace{-5 mm}
\subsection{Mechanical Model Validation}\label{subsec:modelvalidation}
\vspace{-3 mm}
\begin{figure*}
\centerline{\includegraphics[width=5.8in,keepaspectratio]{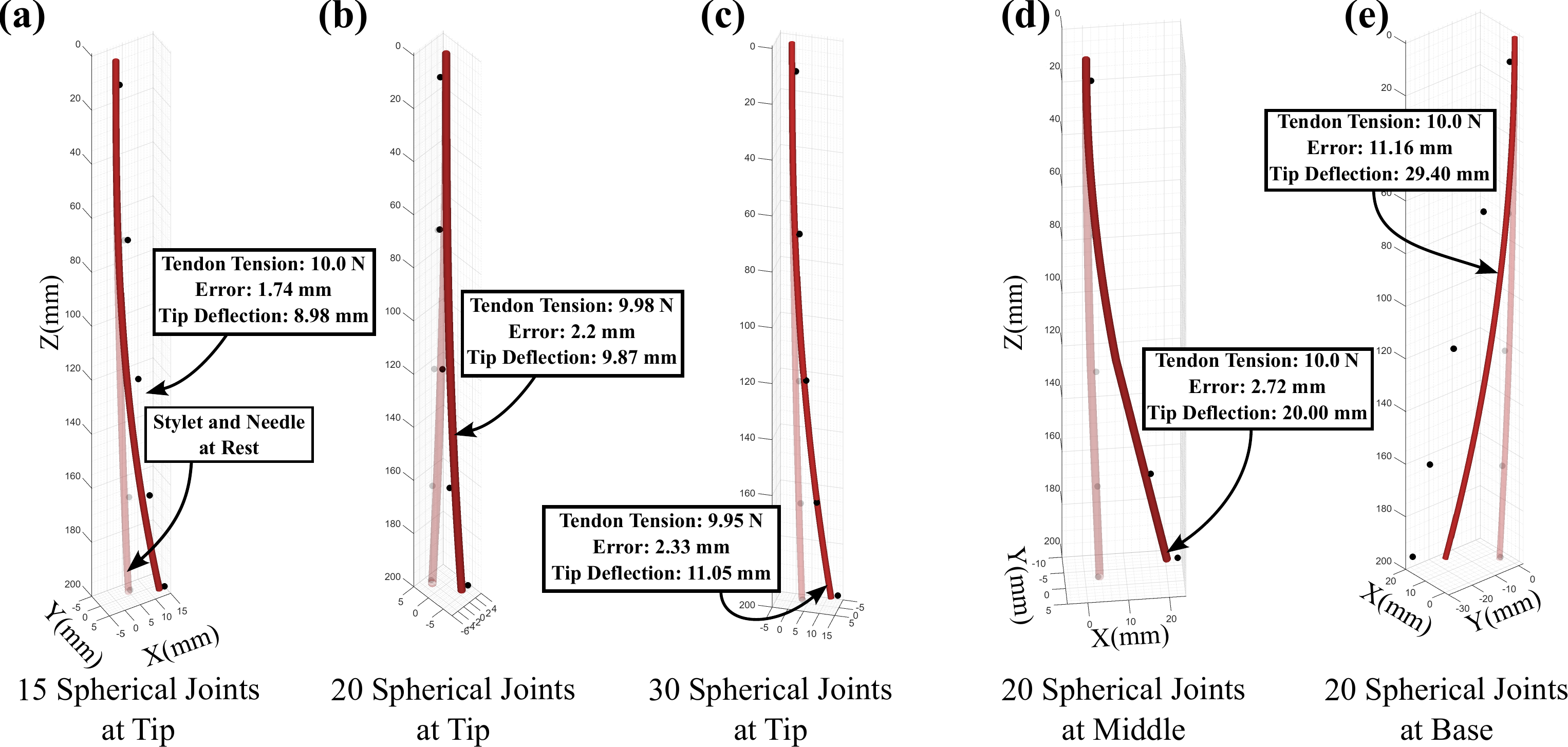}}
\caption{Comparison between the model-predicted centerline (red) and experimental ground-truth data (black markers) for various configurations of the stylet and needle. 
(a-c) depict 15-gauge needle-stylet assembly configurations with varying spherical joint counts. 
(d-e) present configurations with 20 spherical joints while varying placement from the base to the midsection of the stylet.}
\label{fig:ModelvsActual}
\vspace{-3 mm}
\end{figure*}
Figure~\ref{fig:ModelvsActual} illustrates the predicted and experimentally measured shapes of the $15$ gauge needle-stylet configurations with varying spherical joint counts and positions.
For the model, the elastic and shear moduli of the nitinol tube are set to $E_1 = 68~\text{GPa}$ and $G_1 = 25.5~\text{GPa}$ respectively, while values for the needle are $E_2 = 3.2~\text{GPa}$ and $G_2 = 1.06~\text{GPa}$. 
The Euclidean distance between the model-predicted and measured tip positions is calculated for the maximum applied tendon tension in each trial (approximately $10$~N).
When varying the number of spherical joints from $15$ to $30$, the model-predicted tip position error is $1.74$~mm (less than $1\%$ of total length) for $15$ joints, $2.2$~mm ($1.1\%$ of total length) for 20 joints, and $2.33$~mm ($1.1\%$ of total length) for 30 joints. 
Subsequently, the joint position is varied along the stylet, from tip to midsection to base, while maintaining 20 spherical joints.
Although the overall tip deflection increases when the joint section is positioned closer to the base, the model error remains low ($1.36\%$ of total length) for the midsection configuration but increases to $11.16$~mm ($5.6\%$ of total length) when the joints are positioned at the base.
This higher discrepancy likely arises because $K_{bt}$ for the articulated joint section is determined experimentally \textcolor{black}{and held constant. Given the $33$~mm joint section, stiffness variations are negligible at the tip but propagate and amplify error when located at the base. While $K_{bt}$ could be estimated via the Parallel Axis Theorem, internal friction between ball bearings and disks makes this approach unreliable. A more accurate method would involve directly measuring joint stiffness through independent experiments. Current reliance on a constant $K_{bt}$ is a limitation, as it may not fully capture the mechanical behavior of each joint section.
}
\vspace{-6 mm}
\subsection{OncoReach \textcolor{black}{Pilot Study}} \label{preclin}
\vspace{-2 mm}
A clinical expert (N = 1) is recruited for the OncoReach \textcolor{black}{pilot study.} The simulation consists of the proposed clinical use workflow:
\textcolor{black}{
\begin{itemize}[leftmargin=*]
    \item Insert tandem and obturator into uterus using an ultrasound imaging system (SonoSite M-Turbo, FUJIFILM, Bothel, Washington, USA) (Fig. \ref{fig:trial1}(a-c)).
    \item Apply template up to the pelvic phantom (Fig. \ref{fig:trial1}(d)).
    \item Insert straight and steered needles towards the Anterior, Medial, and Posterior targets.
    \item Characterize maximum achievable curvature and evaluate the capacity to reach target areas from restricted, less-invasive entry points.
\end{itemize}
}

\begin{figure*}
\centerline{\includegraphics[width=5.5in,keepaspectratio]{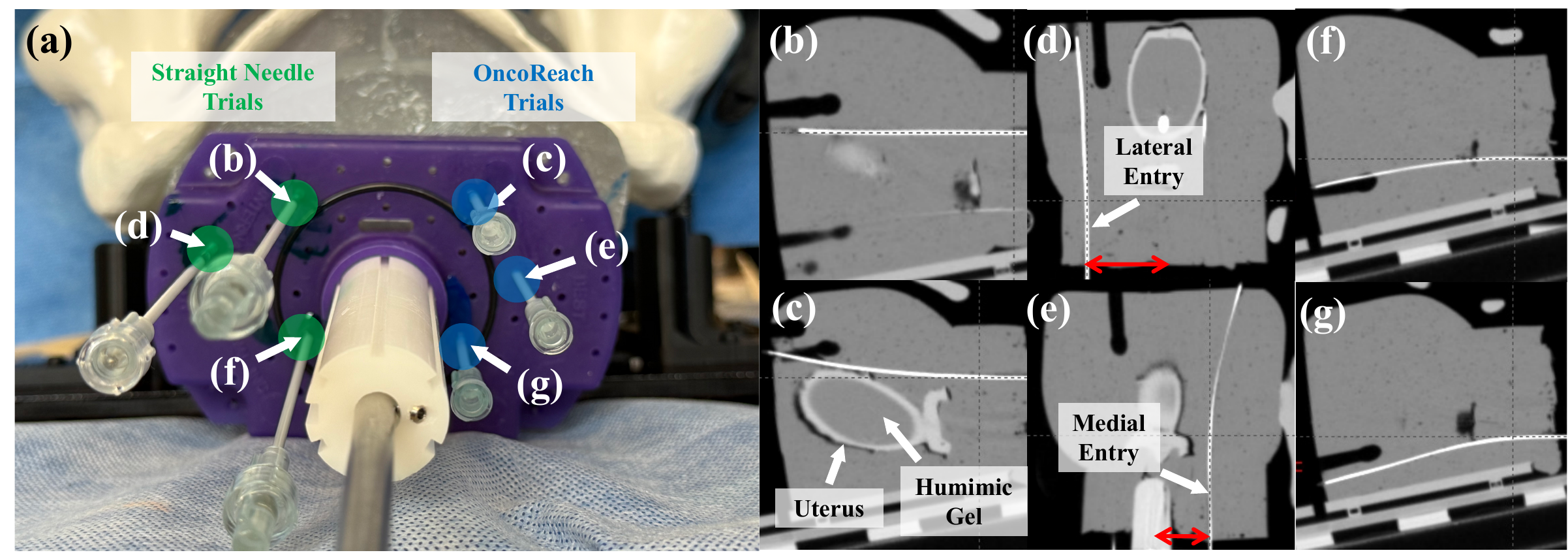}}
\caption{Needle insertion results. (a) Inferior view of the phantom, highlighting entry points and their respective CT sub-figures (target location - straight/steered): (b) Anterior-Straight, (c) Anterior-Steered, (d) Medial-Straight, (e) Medial-Steered, (f) Posterior-Straight, (g) Posterior-Steered.}
\label{fig:trial2}
\vspace{-3 mm}
\end{figure*}

\textcolor{black}{Fig. \ref{fig:trial2}(a) shows the template entry points for both straight needle and OncoReach trials, which are symmetric on opposite sides of the template. Note that for the Medial targets, however, a more central template entry point is chosen for} the OncoReach Trial to demonstrate the ability to steer from a \textcolor{black}{\textit{less-invasive}} entry point (transvaginal entry). The phantom is then imaged using a Brilliance Big Bore CT (Philips, Amsterdam, Netherlands) with a $1$~mm slice thickness, and radiopaque fish wire is inserted into each plastic needle to better visualize the needle paths. 
The CT images were then imported into the BrachyVision treatment planning system (Varian Medical Systems, Palo Alto, CA, USA) for image analysis and measurements.
\begin{figurehere}
\begin{center}
\centerline{\includegraphics[width=2.5in,keepaspectratio]{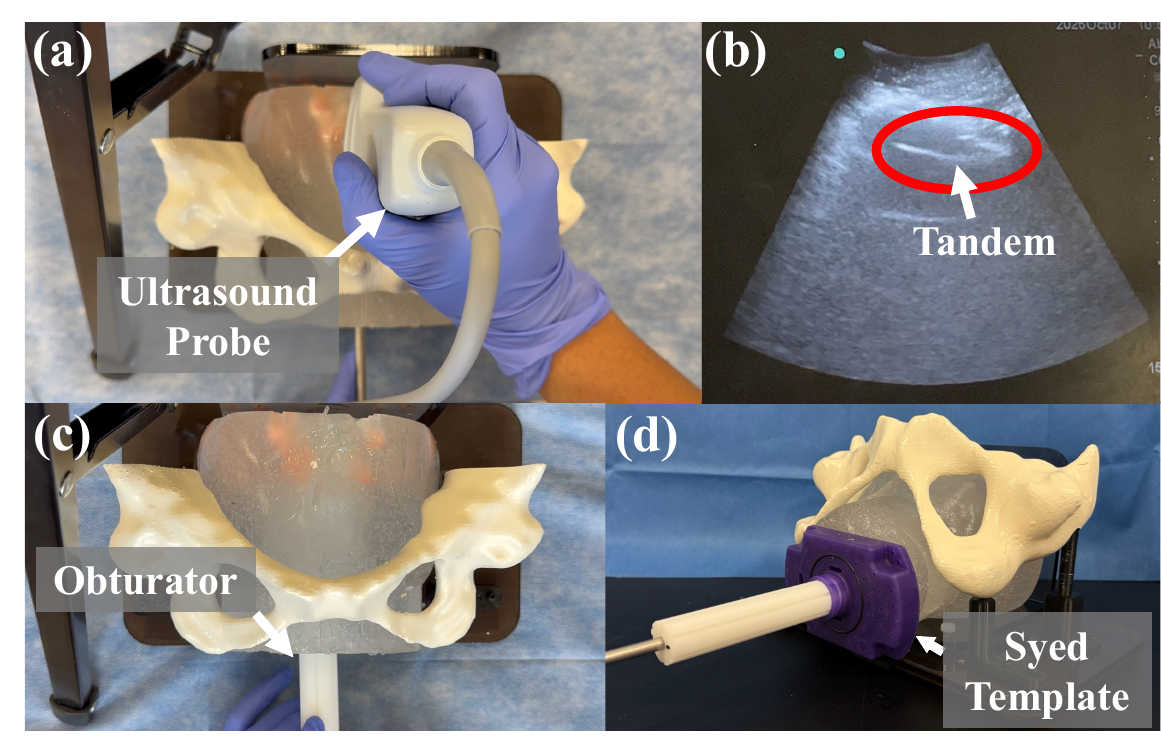}}
\caption{\textcolor{black}{Pilot study setup: phantom with Syed template and tandem attached. }(a) Expert user using the ultrasound probe to visualize tandem inside uterus. (b) Tandem entering the Humimic gel inside the uterus. (c) Obturator fitting inside the vaginal cavity. (d) \textcolor{black}{Pilot study} setup.}
\label{fig:trial1}
\end{center}
\vspace{-5 mm}
\end{figurehere}
The clinical expert can successfully visualize the tandem entering the uterus through the Humimic gel using the ultrasound probe (Fig. \ref{fig:trial1}(a-b)). Both the tandem and vaginal obturator successfully enter the uterus and vaginal cavity (Fig. \ref{fig:trial1}(c)), respectively. The template then successfully slides along the obturator and sits flush on the Humimic pelvic tissue (Fig. \ref{fig:trial1}(d)), showing appropriate anatomical and clinical accuracy in model and model use. However, the pelvis tissue model visually is opaque, preventing users to  assess where the needle is with respect to the target. This can be due to the mold release that mixes with the cast Humimic gel solution while curing.

\vspace{-7 mm}
\begin{tablehere}
\tbl{OncoReach \textcolor{black}{Pilot Study} Results\label{tab:PhantomTrials}}
{\begin{tabular}{@{}lcccr@{}}
\toprule
\multirow{2}{*}{\textbf{Target Type - Needle Type}}   & \textbf{Tip}\\
                                & \textbf{Deflection (mm)}\\ 
\colrule
Anterior-Straight   & N/A \\
Anterior-Steered   & 12.6 \\
Medial-Straight     & 5.8 \\
Medial-Steered      & 15.8 \\
Posterior-Straight  & 13.6\\
Posterior-Steered   & 24.0\\
\botrule
\end{tabular}}
\end{tablehere}
\vspace{-7 mm}
\textcolor{black}{During these insertions, tendon tension is manually modulated by the clinician to adjust the trajectory towards the target region.}
The OncoReach system can properly load and unload a standard 15-gauge brachytherapy needle. The loaded needle can then properly enter the template and the tissue. Fig. \ref{fig:trial2}(a) shows the entry points chosen for both straight and OncoReach-steered needles, and Fig. \ref{fig:trial2}(b-f) shows the CT scans of the resulting needle placement experiments. 
Table \ref{tab:PhantomTrials} reports the tip deflection manually measured in the BrachyVision treatment planning software. 
Note that the Anterior-Straight tip deflection is reported as N/A; the tip deflection was too minimal to measure in the software. Throughout all trials, the OncoReach system is properly bending the needle, allowing for a curved trajectory towards the desired anterior, middle, and posterior targets. In the anterior test to the 5 mm target, the straight needle in Fig. \ref{fig:trial2}(b) enters the pelvic tissue in a linear fashion to reach the target, whereas the steered needle in Fig. \ref{fig:trial2}(c) successfully traversed the most anterior target by 12.6 mm compared to the anterior straight needle. The mid-plane target trials show that, despite entry from a more medial template hole, the OncoReach system is still able to reach the lateral-most target on the mid-plane (Fig. \ref{fig:trial2}(e)). This shows the potential and benefits of steerable stylets, which can allow the radiation oncologist to choose a less invasive approach compared to a more lateral entry point, such as the Medial-Straight trial shown in Fig. \ref{fig:trial2}(d). In the posterior target trials, as the target is hard to see due to the cloudiness of the Humimic gel, both straight and steered needle trials are attempted with the phantom inverted to provide visual feedback. 
\textcolor{black}{The lateral deflections of $12.6-24.0$~mm achieved in this study provide a corrective capability that addresses clinical needs. Studies of conventional gynecologic brachytherapy needle placement report average transverse tip deflections of $3.6\pm2.1$~mm to $7.9\pm3.0$~mm~[\citen{Karius2025}]. By exceeding these typical error magnitudes, the system demonstrates the curvature required to compensate for placement uncertainties encountered in practice.}
The Posterior-Straight trial (Fig. \ref{fig:trial2}(f)) resulted in final needle placement closer to the desired posterior 15 mm target compared to the Posterior-Steered trial ((Fig. \ref{fig:trial2}(g)). This can be attributed to the Syed template sagging, resulting in an angled needle entry and continued trajectory. Additionally, the inability to assess the orientation of the needle tip (i.e. steering direction) once the distal tip entered the tissue is a common observation throughout all trials.
\vspace{-5 mm}
\section{\textcolor{black}{Discussions and Conclusions}} 
\label{sec:conclusion}
\vspace{-2 mm}
In this work, we present the design and \textcolor{black}{pilot study} of a steerable stylet for ISBT applications. Various routing disk designs are experimentally tested in free space, and a mechanical model is developed to predict the robot's shape under applied tendon tension. When the spherical joints are positioned at the tip or midsection, the tip-position error remains below $2\%$ of the needle length. However, when the spherical joints are placed near the base, error propagation along the length increases the tip-position error to $5.6\%$. 
\textcolor{black}{The 15-gauge configuration with 30 spherical joints positioned at the tip was selected for the pilot study. A pilot study was conducted with a single expert user to evaluate the clinical feasibility of the device. The results demonstrated that the handheld device is functional and capable of reaching lateral targets from a safer, medial entry point-a trajectory achievable only through needle steering.}

\vspace{-3 mm}
\subsection{\textcolor{black}{Limitation and Future Work}} \label{sec:Limit}
\vspace{-3 mm}
\textcolor{black}{Although the first pilot study for the OncoReach stylet shows promising translational potential, the current study has several limitations. First, the pilot study consists of a single expert user ($N=1$), which demonstrates the feasibility of the system but does not provide statistical data on generalizable performance or broad clinical usability. A larger user study is required to validate the system's learning curve and consistency across different users.
Second, the current mechanical model is validated only in free space. It does not account for reaction forces from homogeneous or heterogeneous tissue. While the device successfully achieved curvature in the phantom, accurate trajectory prediction in a clinical setting will require integrating tissue-force models and implementing real-time feedback control to compensate for unmodeled disturbances. Finally, due to the opacity of the phantom tissue, real-time visualization of targeting during insertion and quantification of tip position accuracy are not possible.}
\textcolor{black}{Future work will focus on developing a transparent, tissue-mimicking phantom, integrating real-time shape sensing for feedback control, conducting repeatability tests in both free-space and in the phantom model, and conducting a comprehensive preclinical evaluation with a larger group of expert users.}

\vspace{-5 mm}
\nonumsection{Acknowledgments}
\vspace{-1 mm}
\noindent \textcolor{black}{This work was supported in part by a grant from the University of Louisville School of Medicine.}

\bibliographystyle{IEEEtran}
\bibliography{references}

\begin{thebibliography}{10}
\providecommand{\url}[1]{#1}
\csname url@samestyle\endcsname
\providecommand{\newblock}{\relax}
\providecommand{\bibinfo}[2]{#2}
\providecommand{\BIBentrySTDinterwordspacing}{\spaceskip=0pt\relax}
\providecommand{\BIBentryALTinterwordstretchfactor}{4}
\providecommand{\BIBentryALTinterwordspacing}{\spaceskip=\fontdimen2\font plus
\BIBentryALTinterwordstretchfactor\fontdimen3\font minus \fontdimen4\font\relax}
\providecommand{\BIBforeignlanguage}[2]{{%
\expandafter\ifx\csname l@#1\endcsname\relax
\typeout{** WARNING: IEEEtran.bst: No hyphenation pattern has been}%
\typeout{** loaded for the language `#1'. Using the pattern for}%
\typeout{** the default language instead.}%
\else
\language=\csname l@#1\endcsname
\fi
#2}}
\providecommand{\BIBdecl}{\relax}
\BIBdecl

\bibitem{jama}
\BIBentryALTinterwordspacing
Z.~Shahmoradi, H.~Damgacioglu, M.~A. Clarke, N.~Wentzensen, J.~Montealegre, K.~Sonawane, and A.~A. Deshmukh, ``Cervical cancer incidence among us women, 2001-2019,'' \emph{JAMA}, vol. 328, no.~22, pp. 2267--2269, 12 2022. [Online]. Available: \url{https://doi.org/10.1001/jama.2022.17806}
\BIBentrySTDinterwordspacing

\bibitem{Siegel2025}
\BIBentryALTinterwordspacing
R.~L. Siegel, T.~B. Kratzer, A.~N. Giaquinto, H.~Sung, and A.~Jemal, ``Cancer statistics, 2025,'' \emph{CA: A Cancer Journal for Clinicians}, vol.~75, no.~1, pp. 10--45, 2025. [Online]. Available: \url{https://acsjournals.onlinelibrary.wiley.com/doi/abs/10.3322/caac.21871}
\BIBentrySTDinterwordspacing

\bibitem{Brachytherapy_An_overview}
\BIBentryALTinterwordspacing
C.~Chargari, E.~Deutsch, P.~Blanchard, S.~Gouy, H.~Martelli, F.~Guérin, I.~Dumas, A.~Bossi, P.~Morice, A.~N. Viswanathan, and C.~Haie-Meder, ``Brachytherapy: An overview for clinicians,'' \emph{CA: A Cancer Journal for Clinicians}, vol.~69, no.~5, pp. 386--401, 2019. [Online]. Available: \url{https://acsjournals.onlinelibrary.wiley.com/doi/abs/10.3322/caac.21578}
\BIBentrySTDinterwordspacing

\bibitem{Imaging_of_implant_needles}
\BIBentryALTinterwordspacing
F.-A. Siebert, M.~Hirt, P.~Niehoff, and G.~Kovács, ``Imaging of implant needles for real-time hdr-brachytherapy prostate treatment using biplane ultrasound transducers,'' \emph{Medical Physics}, vol.~36, no.~8, pp. 3406--3412, 2009. [Online]. Available: \url{https://aapm.onlinelibrary.wiley.com/doi/abs/10.1118/1.3157107}
\BIBentrySTDinterwordspacing

\bibitem{Anteverted_Uterus}
``\BIBforeignlanguage{en}{Anteverted uterus},'' \url{https://my.clevelandclinic.org/health/diseases/22569-anteverted-uterus}, Mar. 2022, accessed: 2025-10-27.

\bibitem{Millard2024_qj}
E.~Millard, ``\BIBforeignlanguage{en}{What is an anteverted uterus?}'' \url{https://www.healthcentral.com/womens-health/anteverted-uterus}, May 2024, accessed: 2025-10-27.

\bibitem{Stock1997}
\BIBentryALTinterwordspacing
R.~G. Stock, K.~Chan, M.~Terk, J.~Dewyngaert, N.~N. Stone, and P.~Dottino, ``A new technique for performing syed-neblett template interstitial implants for gynecologic malignancies using transrectal-ultrasound guidance,'' \emph{International Journal of Radiation Oncology, Biology, Physics}, vol.~37, no.~4, pp. 819--825, Mar 1997. [Online]. Available: \url{https://doi.org/10.1016/S0360-3016(96)00558-5}
\BIBentrySTDinterwordspacing

\bibitem{TAMS2024}
P.~Kheradmand, B.~Moradkhani, H.~Jella, K.~Sowards, S.~R. Silva, and Y.~Chitalia, ``Towards a tendon-assisted magnetically steered (tams) robotic stylet for brachytherapy,'' \emph{IEEE Robotics and Automation Letters}, vol.~9, no.~7, pp. 6464--6471, 2024.

\bibitem{Gunderman2022}
A.~L. Gunderman, E.~J. Schmidt, M.~Morcos, J.~Tokuda, R.~T. Seethamraju, H.~R. Halperin, A.~N. Viswanathan, and Y.~Chen, ``Mr-tracked deflectable stylet for gynecologic brachytherapy,'' \emph{IEEE/ASME Transactions on Mechatronics}, vol.~27, no.~1, pp. 407--417, 2022.

\bibitem{Deaton2023SteerableStylet}
N.~J. Deaton \emph{et~al.}, ``Towards steering a high-dose rate brachytherapy needle with a robotic steerable stylet,'' \emph{IEEE Transactions on Medical Robotics and Bionics}, vol.~5, no.~1, pp. 54--65, Feb 2023.

\bibitem{deaton2021robotically}
N.~J. Deaton, Y.~Chitalia, P.~Patel, and J.~P. Desai, ``Towards a robotically steerable system for high dose rate brachytherapy,'' in \emph{Proceedings of the Conference on Experimental Robotics}, 2021, pp. 233--244.

\bibitem{Deaton2023}
N.~J. Deaton, M.~Sheft, and J.~P. Desai, ``Towards fbg-based shape sensing and sensor drift for a steerable needle,'' \emph{IEEE/ASME Transactions on Mechatronics}, vol.~28, no.~6, pp. 3041--3052, Dec 2023.

\bibitem{CalebStatic2011}
D.~C. Rucker and R.~J. Webster~III, ``Statics and dynamics of continuum robots with general tendon routing and external loading,'' \emph{IEEE Transactions on Robotics}, vol.~27, no.~6, pp. 1033--1044, 2011.

\bibitem{dupont2009design}
P.~E. Dupont, J.~Lock, B.~Itkowitz, and E.~Butler, ``Design and control of concentric-tube robots,'' \emph{IEEE Transactions on Robotics}, vol.~26, no.~2, pp. 209--225, 2010.

\bibitem{gazzola2018forward}
M.~Gazzola, L.~Dudte, A.~McCormick, and L.~Mahadevan, ``Forward and inverse problems in the mechanics of soft filaments,'' \emph{Royal Society open science}, vol.~5, no.~6, p. 171628, 2018.

\bibitem{chitalia2023model}
Y.~Chitalia, A.~Sarma, T.~A. Brumfiel, N.~J. Deaton, M.~Sheft, and J.~P. Desai, ``Model-based design of the coast guidewire robot for large deflection,'' \emph{IEEE robotics and automation letters}, vol.~8, no.~9, pp. 5345--5352, 2023.

\bibitem{julie}
B.~Eckroate, D.~Ayala-Peacock, R.~Venkataraman, S.~Campelo, J.~Chino, S.~J. Stephens, Y.~Kim, S.~Meltsner, J.~Raffi, and O.~Craciunescu, ``A novel multi-modality imaging phantom for validating interstitial needle guidance for high dose rate gynecological brachytherapy,'' \emph{Journal of Applied Clinical Medical Physics}, vol.~24, no.~10, p. e14075, 2023.

\bibitem{model2}
L.~H. Bloom, D.~Ayala-Peacock, R.~Venkataraman, B.~Eckroate, R.~Sanford, J.~Chino, Y.~Kim, J.~Raffi, and O.~Crăciunescu, ``Implementation of needle‐tracking technology for real‐time transrectal ultrasound‐guided interstitial gynecological hdr brachytherapy: A feasibility study,'' \emph{Journal of Applied Clinical Medical Physics}, vol.~26, no.~6, p. e70100, 2025.

\bibitem{model1}
S.~Campelo, E.~Subashi, S.~G. Meltsner, Z.~Chang, J.~Chino, and O.~Crăciunescu, ``Multimaterial three-dimensional printing in brachytherapy: Prototyping teaching tools for interstitial and intracavitary procedures in cervical cancers,'' \emph{Brachytherapy}, vol.~19, no.~6, pp. 767--776, 2020.

\bibitem{3dslicer}
A.~Fedorov, R.~Beichel, J.~Kalpathy-Cramer, J.~Finet, J.-C. Fillion-Robin, S.~Pujol, C.~Bauer, D.~Jennings, F.~Fennessy, M.~Sonka, J.~Buatti, S.~Aylward, J.~V.~Miller, S.~Pieper, and R.~Kikinis, ``3d slicer as an image computing platform for the quantitative imaging network,'' \emph{Magnetic Resonance Imaging}, vol.~30, no.~9, pp. 1323--1341, 2012, quantitative Imaging in Cancer.

\bibitem{humimic}
{Humimic Medical}, \emph{Humimic SimuGel™ Technical Documentation}, Humimic Medical, n.d., technical specification and acoustic data sheet for SimuGel™ formulations 0--5.

\bibitem{Karius2025}
\BIBentryALTinterwordspacing
A.~Karius, V.~Strnad, M.~Lotter, S.~Kreppner, R.~Merten, R.~Fietkau, C.~Bert, and C.~Schweizer, ``Assessment of needle bending and tracking requirements for optimized needle placement in combined intracavitary/interstitial gynecologic brachytherapy,'' \emph{Strahlentherapie und Onkologie}, Feb 2025. [Online]. Available: \url{https://doi.org/10.1007/s00066-025-02367-2}
\BIBentrySTDinterwordspacing

\end{thebibliography}

\noindent\includegraphics[width=1in]{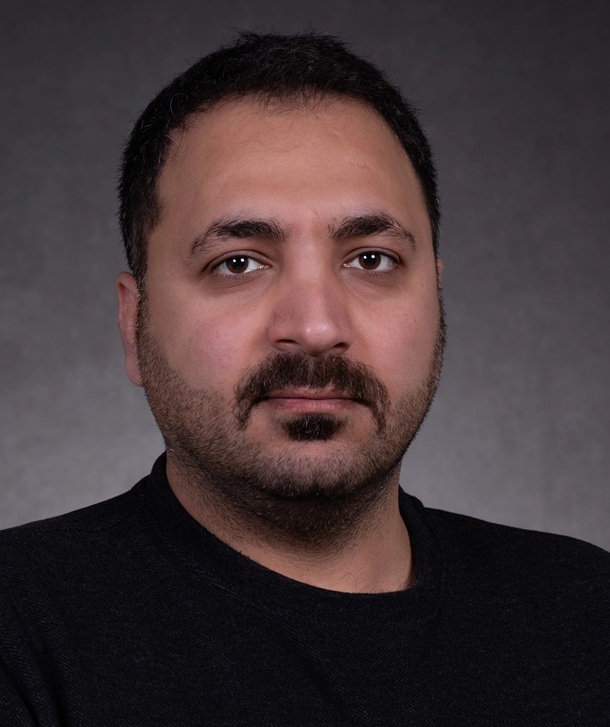}
{\bf Pejman Kheradmand} received the bachelor’s degree in mechanical engineering from Razi University, Kermanshah, Iran, and the master’s degree in mechanical engineering from the University of Tehran, Tehran, Iran. He is currently working toward the Ph.D. degree in mechanical engineering with the University of Louisville, Louisville, KY, USA. His research focuses on continuum robots in surgical contexts.\\

\noindent\includegraphics[width=1in]{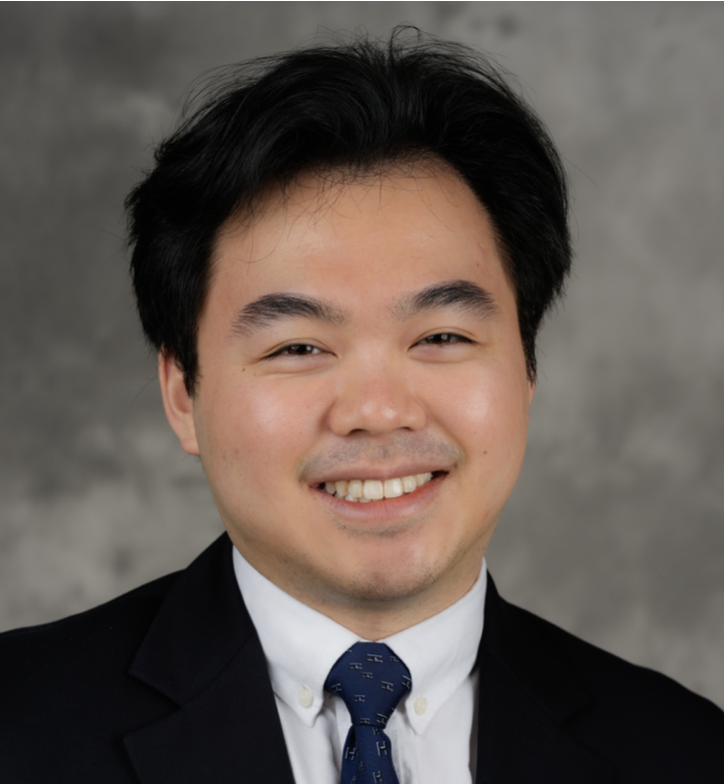}{\bf Kent K. Yamamoto} received his BS in Biomedical Engineering at the Georgia Institute of Technology in 2021. He is currently pursuing his Ph.D. in Mechanical Engineering at Duke University co-advised by Dr. Patrick Codd and Dr. Yash Chitalia. He also serves as the Engineering Coordinator at the Duke University School of Medicine Surgical Education and Activities Lab (SEAL). His research interests include mechanical design, surgical laser steering, and continuum robotics for minimally invasive surgery.\\

\noindent\includegraphics[width=1in]{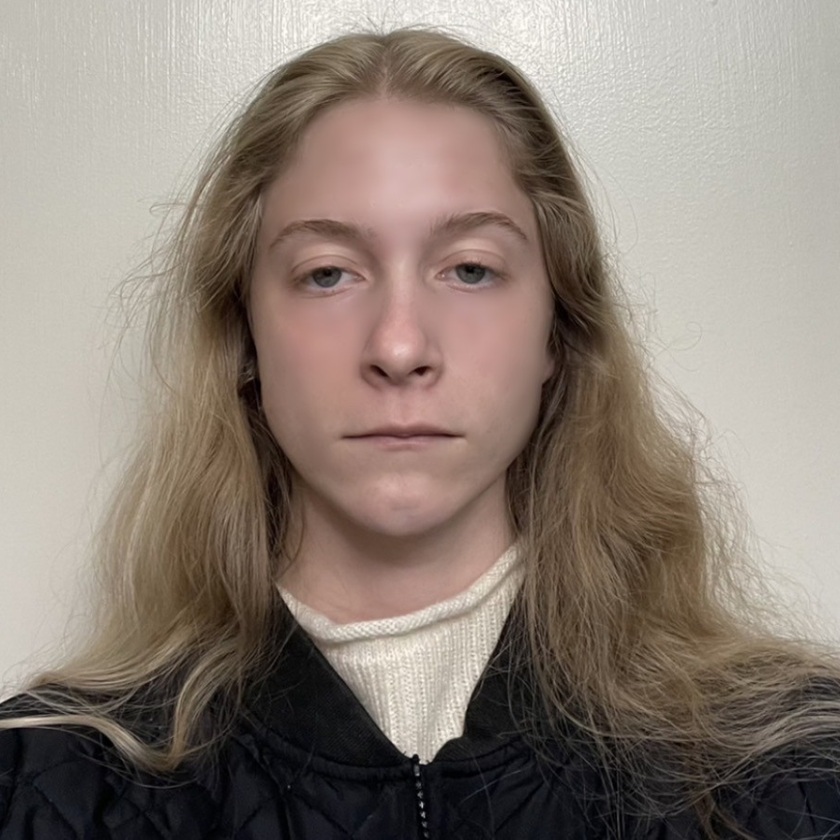}{\bf Emma Webster} is an undergraduate mechanical engineering student at the University of Louisville, pursuing a minor Philosophy.\\

\noindent\includegraphics[width=1in]{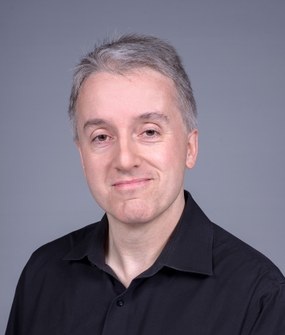}{\bf Keith Sowards} received his B.S. in Physics from the University of Kentucky in 1995. He earned his M.S. in Health Physics from the University of Kentucky in 1999. He is currently a Clinical Physicist in the Department of Radiation Oncology at the University of Louisville.\\

\noindent\includegraphics[width=1in]{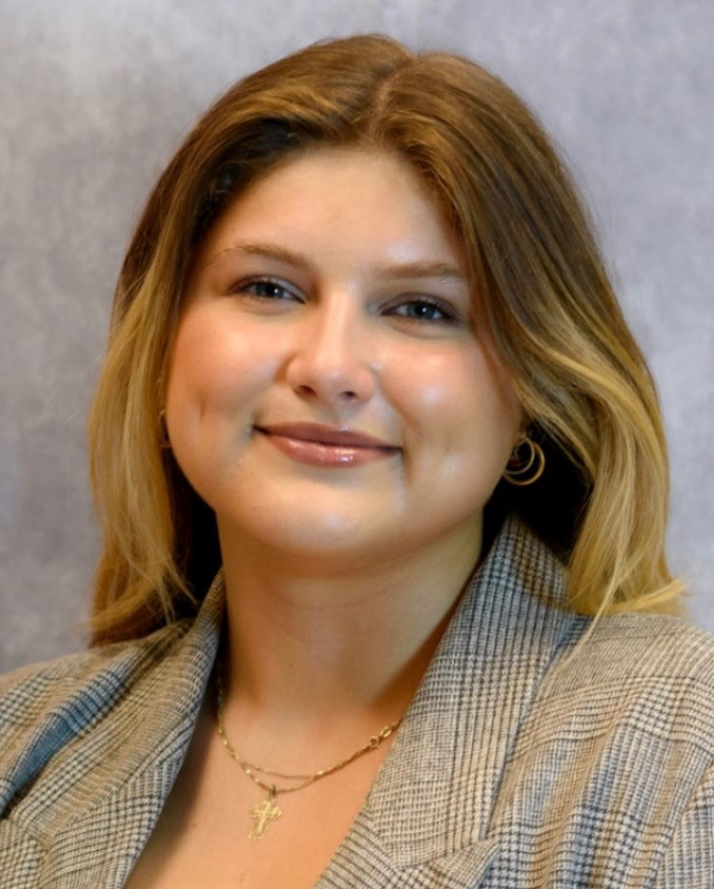}{\bf Gianna Hatheway} received her B.S. in Biological Engineering from The Ohio State University in 2022. She is currently pursuing her Master’s Degree in Medical Physics at Duke University, co-advised by Dr. Julie Raffi and Dr. Yang Sheng. Her research interests include the use of artificial intelligence and automation for brachytherapy treatment planning. \\

\noindent\includegraphics[width=1in]{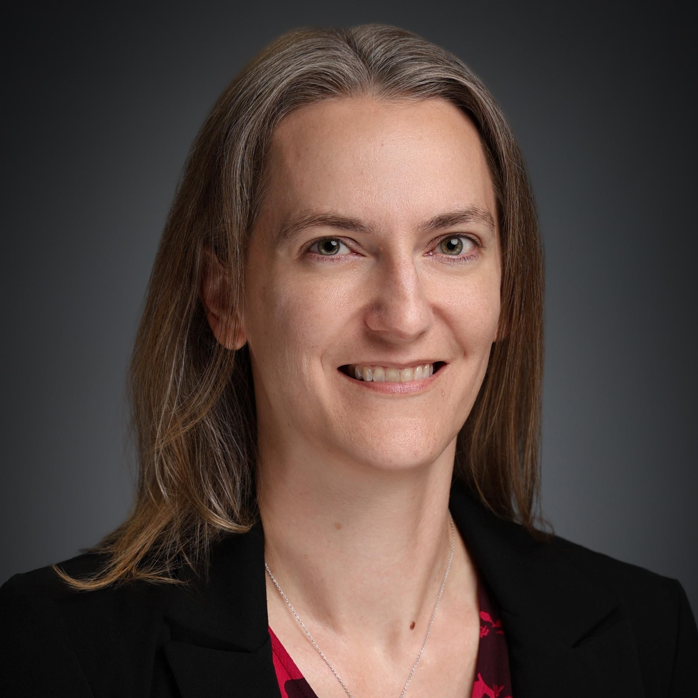}{\bf Julie A. Raffi} received her B.A. degree in Physics from the College of St. Benedict in 2004. She earned her M.S. and Ph.D. degrees in Medical Physics from the University of Wisconsin – Madison in 2007 and 2010, respectively. She is currently an Assistant Professor in the Department of Radiation Oncology at Duke University and an associate faculty member in the Duke Medical Physics Graduate Program. Her research interests include developing anatomical phantoms with 3D printing and molding techniques and adapting existing and emerging needle guidance technology and optimization strategies for gynecologic brachytherapy applications.

\noindent\includegraphics[width=1in]{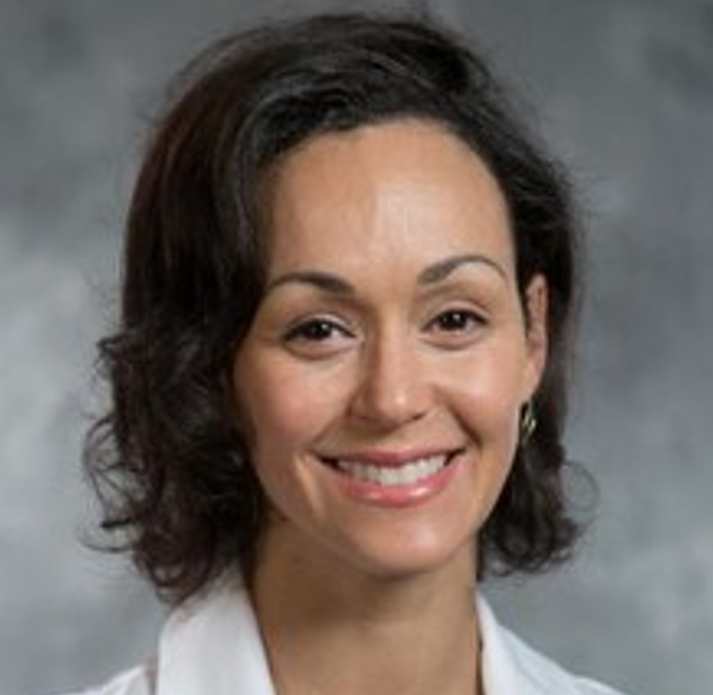}{\bf Diandra N. Ayala-Peacock} received her B.S. in Molecular Biophysics and Biochemistry from Yale University. She received her MD degree and completed both internship and subsequent Radiation Oncology residency and chief year at Wake Forest University Medical Center. She is now an Associate Professor of Radiation Oncology at Duke University. Dr. Ayala-Peacock's research interests include advancement of brachytherapy-focused technologies including imaging fusion, guidance and tracking, as well as pelvic phantom development for evaluation of these technologies. She is also active in national clinical trial development for gynecologic malignancies where she holds several leadership positions and has additional roles in several institutional studies regarding sexual health concerns among cancer survivors.\\

\noindent\includegraphics[width=1in]{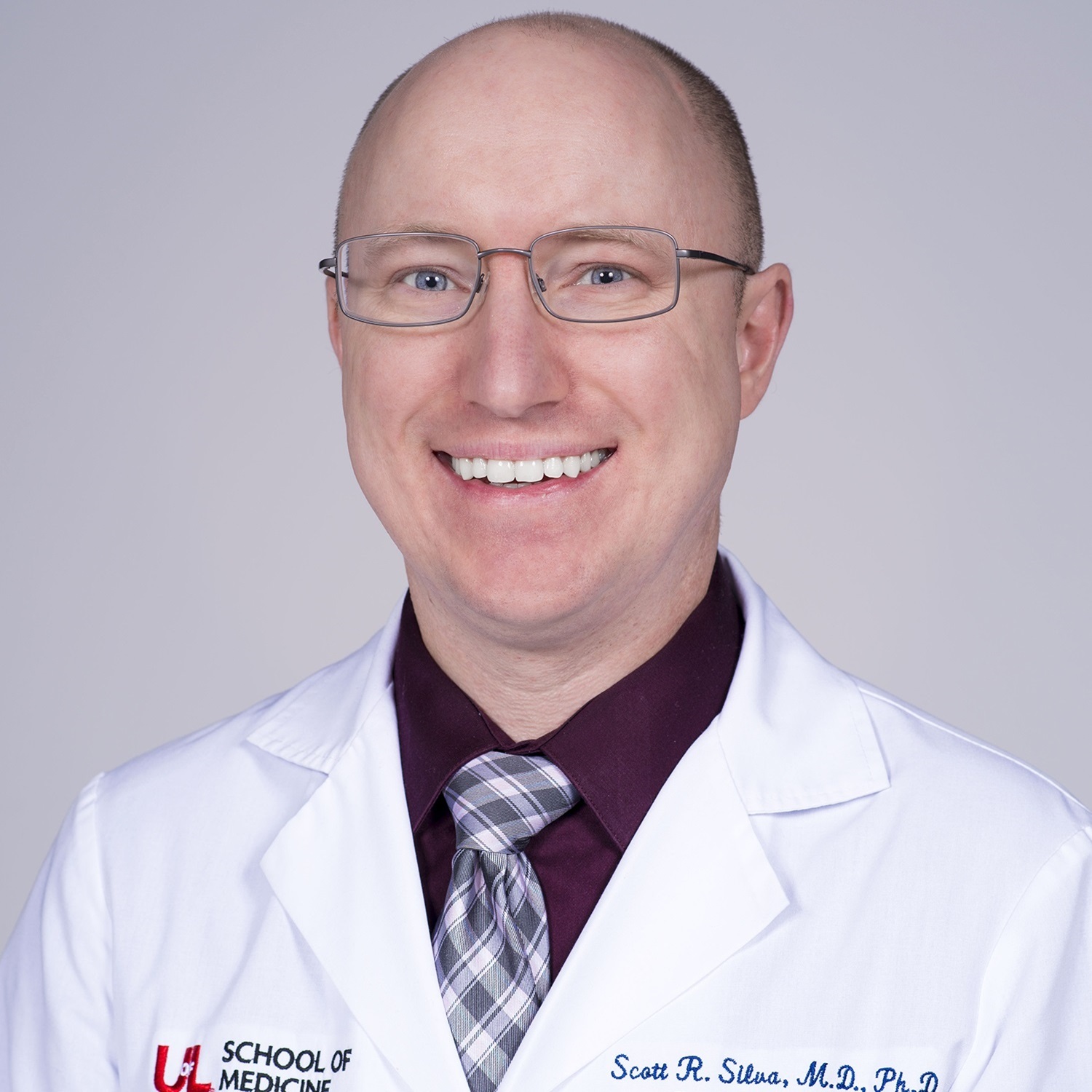}
{\bf Scott R. Silva}  received his B.S. in Biochemistry from the University of Texas at Austin and his Ph.D. in Molecular and Cellular Biology from the University of Kentucky. He earned his M.D. from the University of Kentucky College of Medicine. He completed his General Surgery Internship at Vanderbilt University and his Residency in Radiation Oncology at Loyola University, where he also served as Chief Resident in 2018. He is currently the Director of Brachytherapy and an Associate Professor in the Department of Radiation Oncology at the University of Louisville, Louisville, Kentucky.\\

\noindent\includegraphics[width=1in]{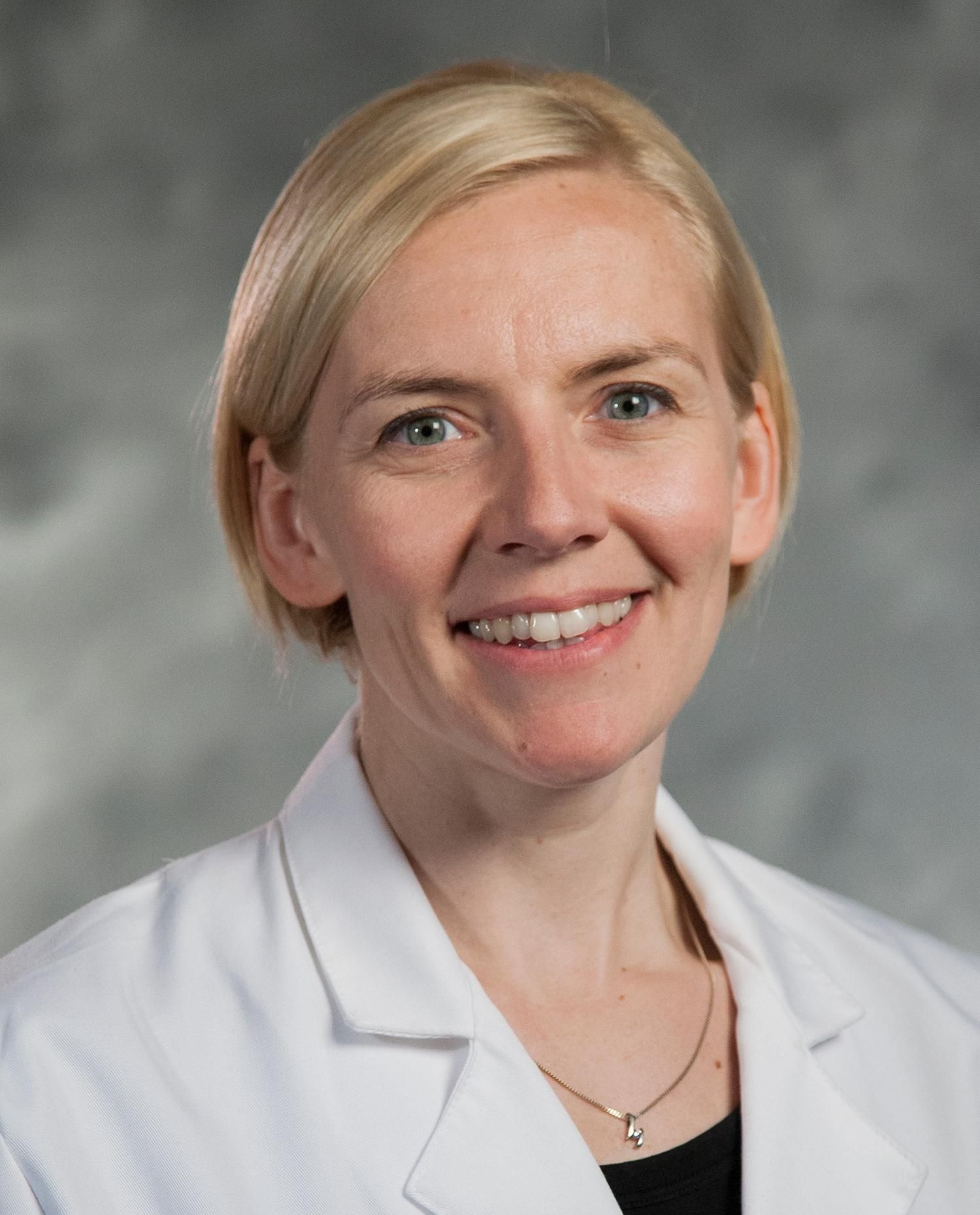}
{\bf Katharine L. Jackson} received her M.D. from  Oxford University and University College London. She graduated from a general surgery residency at Brown University and the University of Utah, followed by a fellowship in Colorectal Surgery at the Cleveland Clinic Foundation, Ohio from which she graduated in 2013. Dr Jackson is an Assistant Professor in the Department of Surgery, Section of Colorectal Surgery, the Education Director of the Surgical Education and Activities Lab (SEAL), and the Head of Robotic Surgery at Duke Raleigh Hospital. She has a passion for developing engaging strategies for surgical education and implementing innovative strategies and technologies that improve patient care in a sustainable way.

\noindent\includegraphics[width=1in]{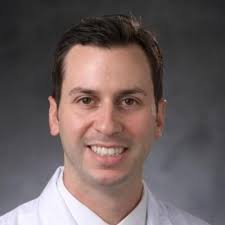}
{\bf Sabino Zani Jr.} received his M.D. from Albany Medical College, completed his General Surgery residency at The University of Connecticut Integrative, and completed his Surgical Oncology Fellowship at Duke University. He is now an Associate Professor of Surgery and the Mechanical Engineering and Materials Science (MEMS) Department at Duke University and the Co-Director of the Surgical Education and Activities Lab (SEAL).

\noindent\includegraphics[width=1in]{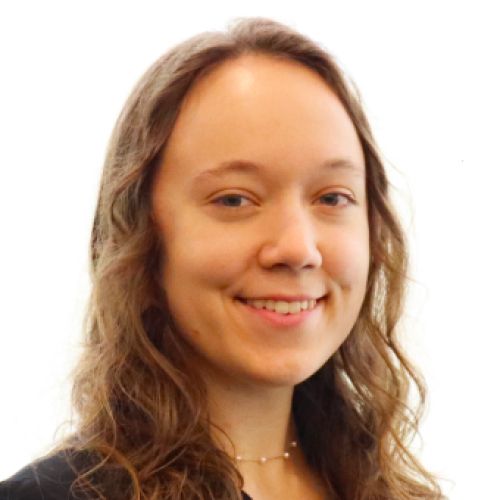}
{\bf Joanna Deaton Bertram} received her B.S. degree in Biomedical Engineering in 2018 from the Georgia Institute of Technology, Atlanta, GA, USA, where she also earned her Ph.D. degree in Robotics in 2024. She is currently an Assistant Professor in the Department of Mechanical Engineering and Materials Science at Duke University, where she leads research in medical robotics. Her work focuses on advancing the design, modeling, and control of robotic systems for medical applications, with particular emphasis on continuum robotics, image and sensor feedback, integrated sensor design, smart materials, and surgical robotics.\\

\noindent\includegraphics[width=1in]{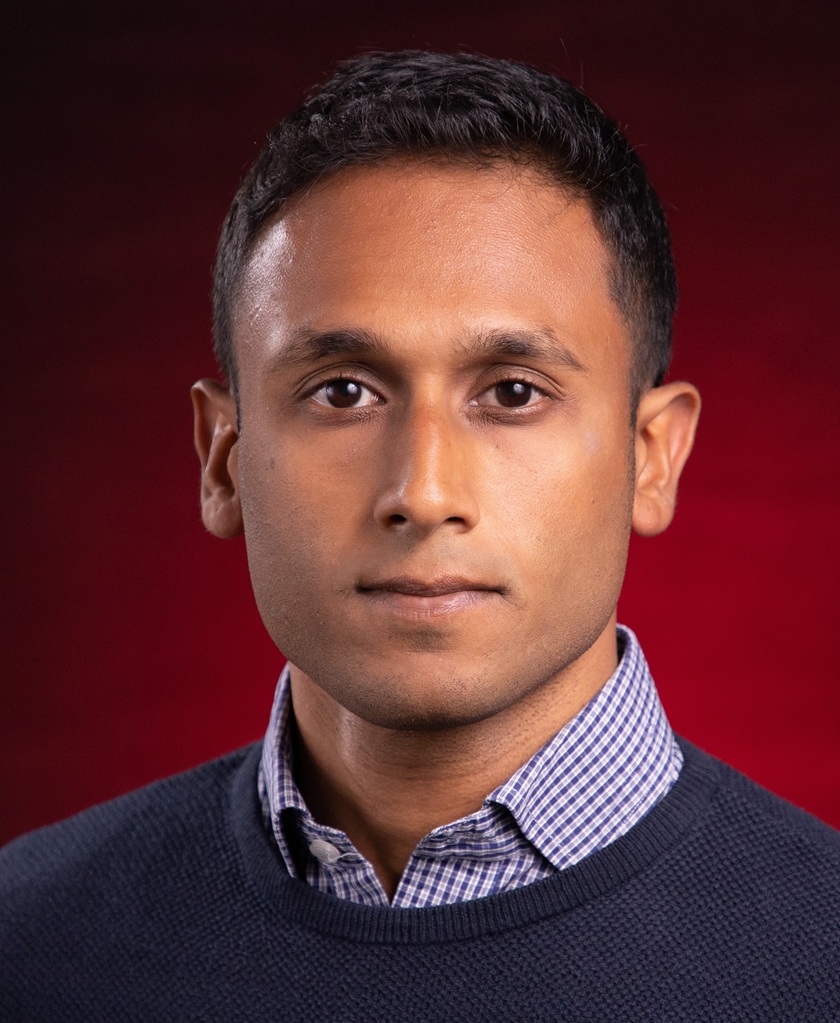}
{\bf Yash Chitalia} (Member, IEEE) received the bachelor’s degree in electronics engineering from the University of Mumbai, Mumbai, India, in 2011, the M.S. degree in electrical engineering from the University of Michigan, Ann Arbor, MI, USA, in 2013, and the Ph.D. degree in mechanical engineering from the Georgia Institute of Technology, Atlanta, GA, USA, in 2021. He is currently an Assistant Professor with the Department of Mechanical Engineering, University of Louisville, Louisville, KY, USA. His research interests include microscale and mesoscale surgical robots for cardiovascular and neurosurgical applications as well as rehabilitation robotics. Dr. Chitalia was the recipient of several awards including the Ralph E. Powe Junior Faculty Enhancement Award and the NASA-KY Research Initiation Award.

\end{multicols}

\end{document}